\documentclass[journal]{IEEEtran} 

\usepackage{cite}                   

\usepackage{graphicx}               
\DeclareGraphicsExtensions{.pdf}    

\usepackage{url}                    

\usepackage[utf8]{inputenc}         
\usepackage{amsmath}                
\usepackage{amsfonts}               
\usepackage{amssymb}
\usepackage{bm}                     

\usepackage{algorithm}              
\usepackage{algpseudocode}          
\newcommand{\algofontsize}{\small}  

\usepackage{tabularx}               
\usepackage{booktabs}               
\usepackage{multirow}               
\usepackage{dcolumn}                
\usepackage{makecell}               

\usepackage{caption}                
\begin{document}

\title{Swarm Characteristic Classification using\\
        Robust Neural Networks with\\
        Optimized Controllable Inputs}
\author{Donald W. Peltier III,
        Isaac Kaminer,
        Abram Clark,
        Marko Orescanin%
\thanks{Authors are with the Naval Postgraduate School, Monterey, CA, 93943.}%
\thanks{D. Peltier is with the Department of Mechanical \& Aerospace Engineering.}%
\thanks{I. Kaminer is with the Department of Mechanical \& Aerospace Engineering.}%
\thanks{A. Clark is with the Department of Physics.}%
\thanks{M. Orescanin is with the Department of Computer Science.}}%
\maketitle

\begin{abstract}
Having the ability to infer characteristics of autonomous agents would profoundly revolutionize defense, security, and civil applications. Our previous work was the first to demonstrate that supervised neural network time series classification (NN TSC) could rapidly predict the tactics of swarming autonomous agents in military contexts, providing intelligence to inform counter-maneuvers. However, most autonomous interactions, especially military engagements, are fraught with uncertainty, raising questions about the practicality of using a pretrained classifier. This article addresses that challenge by leveraging expected operational variations to construct a richer dataset, resulting in a more robust NN with improved inference performance in scenarios characterized by significant uncertainties. Specifically, diverse datasets are created by simulating variations in defender numbers, defender motions, and measurement noise levels. Key findings indicate that robust NNs trained on an enriched dataset exhibit enhanced classification accuracy and offer operational flexibility, such as reducing resources required and offering adherence to trajectory constraints. Furthermore, we present a new framework for optimally deploying a trained NN by the defenders. The framework involves optimizing defender trajectories that elicit  adversary responses that maximize the probability of correct NN tactic classification while also satisfying operational constraints imposed on the defenders.
\end{abstract}

\begin{IEEEkeywords}
swarms, autonomous systems, neural networks, data augmentation, robustness, onboard optimization, fleet coordination, autonomous agent interactions, optimal control, optimized neural network input
\end{IEEEkeywords}

\section{Introduction}
\label{sec:intro}

\IEEEPARstart{T}{he} relevance of defense systems designed to mitigate attacks by autonomous agents has grown significantly in recent years, particularly in the context of modern conflicts. These engagements have highlighted the staggering effectiveness of autonomous systems, such as aerial drones, bomb-laden boats, and glide bombs infused with artificial intelligence \cite{balmforth_ukrainian_2023, robins-early_ais_2024}. Now consider future attacks using these same systems but at a greater scale; the ability of such swarms to saturate defensive measures presents a formidable challenge to conventional defense strategies \cite{kallenborn_are_2020, pettyjohn_swarms_2024, cummings_artificial_2017}. Defense departments across multiple governments have identified swarm defense as a critical focus area, emphasizing the urgent need for advanced techniques to predict and counteract the tactics and capabilities of these swarms \cite{clark_hicks_2023, hepworth_report_2022}.

Beyond military applications, the ability to predict the characteristics of agents in a ``swarm" has broad implications for civil use cases as well. In scenarios where perception, enabled by exteroceptive sensors, can be used to infer the capabilities, limitations, and intent of non-communicative actors, such predictive techniques are highly valuable. For instance, in autonomous driving, understanding the behavior of other vehicles at intersections, during merging, or for collision avoidance is crucial for safety \cite{lin_ethics_2013, jafary_survey_2018}. Similarly, in regulated airspace and congested waterways, ensuring efficient vehicle routing and preventing collisions is paramount \cite{emha_abdillah_implementation_2024, soori_artificial_2023, vagale_path_2021}. In automated workspaces, such as warehouses employing autonomous robots and human workers, the ability to predict movements can significantly enhance routing efficiency and reduce operational risks \cite{wurman_coordinating_2008, sanchez_ibanez_path_2021, amoo_ai-driven_2024}.

In our previous work we have addressed the problem of  defending a High Value Unit (HVU) against a swarm attack. In particular, we have shown that the problem of optimal HVU defense can be cast as an optimal control problem using so-called attrition rate functions to model weapon effectiveness of the adversaries and defenders \cite{walton_optimal_2018}. We have also studied a scenario where the adversaries are using potential based algorithms with collision avoidance properties to avoid but not fire at the defenders, thus saving their ammo to destroy the HVU \cite{walton_defense_2022}. The case where the adversaries are also shooting at the defenders is addressed in \cite{tsatsanifos_modeling_2021}. Finally, in \cite{gong_partial_2020} we addressed the problem of estimating coefficients of a potential based algorithm used by the adversaries. We note that the work reported in \cite{walton_optimal_2018,walton_defense_2022,tsatsanifos_modeling_2021,gong_partial_2020} assumed knowledge of the algorithm(s) used by the adversaries. This assumption was first lifted in our paper  \cite{peltier_swarm_2024}, where we have shown that it is indeed possible to use a neural network (NN) classifier to determine the algorithms/tactics used by the attacking swarm.

Building upon our previous work, the specific goals of this paper are twofold. First, we aim to define a process for enriching the swarm dataset from our previous work to ensure robust NN performance under operational uncertainty \cite{peltier_swarm_2024}. Our data enrichment approach draws inspiration from the well-established concept of data augmentation, wherein the original dataset is transformed into multiple variations that are pragmatically selected based on expected operational changes \cite{gao_data_2024, iglesias_data_2023}. In our context, the operational variations explored include controllable degrees of freedom (DOF) such as defender number and defender motion, as well as uncontrollable factors like measurement noise.

The second goal of this paper is to develop a user-friendly framework for optimally deploying a trained NN. We show that the performance of a trained NN classifier can be improved when the defender motion can be designed to optimally influence adversary motion, which in turn affects NN predictions, providing a controllable link between NN inputs and outputs. Ultimately, this  framework aims to produce operationally constrained optimized defender motion and provide insights into the minimum number of defenders required to achieve a desired classification confidence level.

To summarize, the key contributions of this study are: (1) systematically selecting dataset variables for augmentation to build enriched swarm datasets for training robust NNs, (2) analyzing robust NN performance under diverse operational conditions, and (3) developing a framework for optimally deploying a trained NN.

\section{Methodology}

\subsection{Scenario}

This study builds on the scenario and assumptions established in our previous work where a swarm-on-swarm engagement is simulated to generate labeled data for supervised neural network time series classification (NN TSC) \cite{peltier_swarm_2024}. In this scenario, a swarm of weaponless defenders aims to motivate an attacking adversarial swarm to maneuver, thereby revealing the adversary's tactics. The adversaries employ one of four predefined tactics (Greedy, Greedy+, Auction, Auction+) and it's assumed that all adversarial agents use identical tactics with no tactic switching during the engagement and that complete adversary trajectories are known. The simulation environment is two-dimensional, with defenders and adversaries characterized by their positions and velocities in two dimensional (2D) space. 

Supervised NN TSC is used to predict the adversarial swarm tactic based on multivariate time series data, where the inputs are the 2D positions and velocities of attacking adversaries. As demonstrated in our original paper, though NN TSC can provide different output types, this paper focuses on multiclass outputs, where the NN outputs a probability vector corresponding to the four different adversarial swarm tactics.

To evaluate NN performance, accuracy is used as the primary metric, given that the possible classifications are equally represented in our datasets. Additionally, normalized error rate, which is accuracy adjusted and normalized using random guess accuracy, is considered as a secondary metric to provide a more nuanced view of model performance under varying noise conditions. Further details regarding the simulation setup, including the specific algorithms used for adversary targeting and the mathematical modeling of adversary motion dynamics, can be found in our original paper \cite{peltier_swarm_2024}.

\subsection{Dataset Enrichment}
\label{sec:dataset enrichment}

\subsubsection{Picking Variables for Augmentation}
The primary question guiding our dataset enrichment quest was how a pretrained classifier would perform if conditions during deployment differed from those during training. In our engagement simulation, numerous variables exist, and individually each can have a wide range of influence on inference performance for a trained NN classifier. These variables include factors related to both the adversaries and defenders, such as the number of agents, velocity and acceleration constraints, and initial formation shape and dispersion. Additionally, variables specific to the adversaries include weapon and sensor ranges, while those specific to the defenders include motion patterns. Lastly, engagement orientation variables such as the starting swarm separation distance and location geometry can play a significant role. As a baseline, the scenario of 10 defenders versus 10 adversaries was adopted.

Of all the variables identified, we selected key \textit{Variables of Interest} (VOI), which were predicted to have high variability during actual deployment and were also expected to have a significant impact on the classifier's performance. These VOI include controllable DOF such as defender number ($N_D$) and defender motion (DM), as well as uncontrollable factors like measurement noise.

The remaining variables that were not considered in this study were grouped into two categories. First, variables expected to yield trends similar to those investigated were omitted for conciseness, such as formation shape, intraswarm dispersion, weapon and sensor range, and velocity and acceleration constraints. Second, certain variables would require implementing additional NN techniques, such as masking and padding to accommodate variations in the number of adversaries and thus NN inputs, or dataset scaling and basis changes to account for differences in initial swarm separation and engagement geometry. These considerations were beyond the scope of the current study but are acknowledged as important factors for future research, specifically for employing the trained classifier in varying real-world scenarios.

\subsubsection{VOI \#1 - Defender Number}
\begin{figure}[h]
    \centering
    \includegraphics[width=\linewidth]{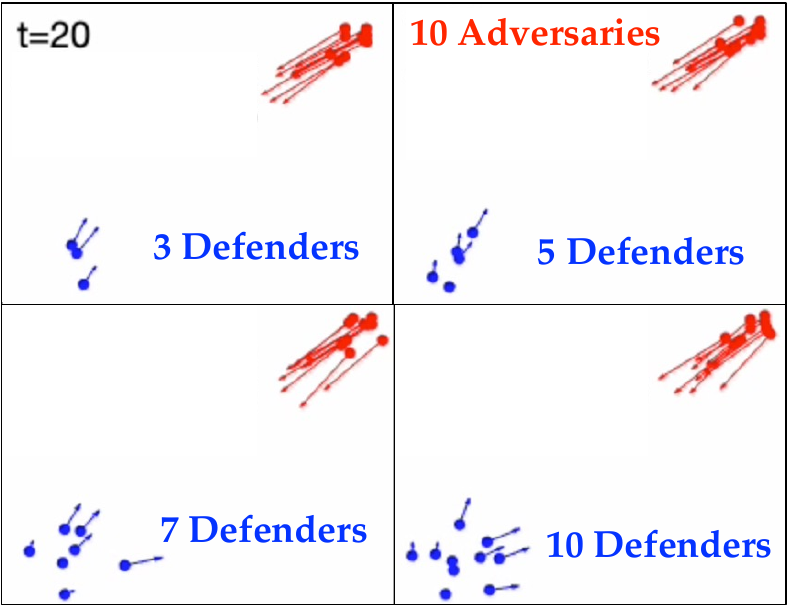}
    \caption{Different adversary responses when varying defender number, even with identical engagement initialization. Velocity vector shown for each agent.}
    \label{fig:dn_each}
\end{figure}

In real-world defensive scenarios, it is highly likely that friendly deployed units will not always be able to match the number of adversaries. This situation can arise due to various constraints, such as surprise attacks, supply chain delays, limited storage facilities, or the availability of only a few nearby collaborative agents. To explore the impact of varying defender number ($N_D$) on NN performance, $N_D$ is varied from 1 to 15, incrementing by 1, resulting in 15 sub-datasets. Of note, $N_D$ was allowed to exceed the number of adversaries (i.e. 10) in order to observe the resultant effects on classifier performance. For visualization, Figure \ref{fig:dn_each} highlights the unique adversary responses (shown in red) at time step 20 based on different $N_D$, proving that even when using identical engagement initialization $N_D$ effects the NN inputs, and hence classification accuracy.

\subsubsection{VOI \#2 - Defender Motion}
\begin{figure}[h]
    \centering
    \includegraphics[width=\linewidth]{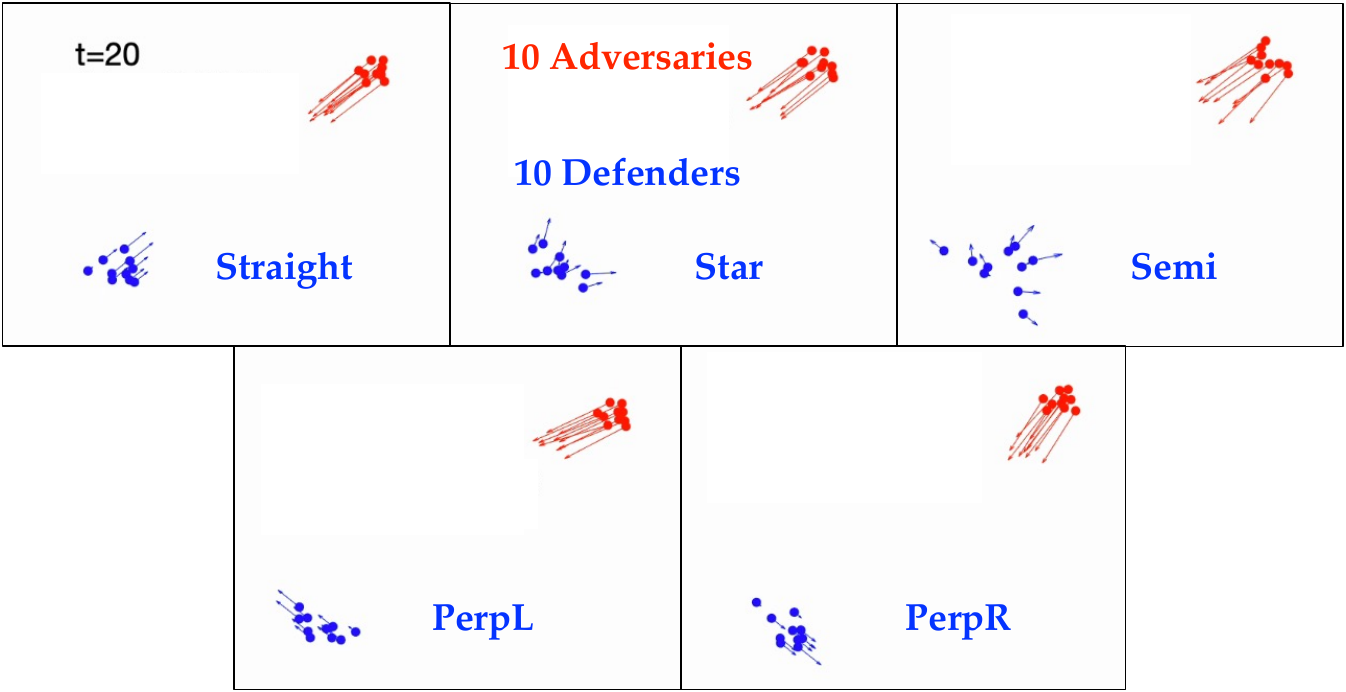}
    \caption{Five distinct defender motions elicit unique adversary responses, even with identical  initialization.}
    \label{fig:dm_each}
\end{figure}

Defender motion is intuitively a controllable DOF that will have one of the strongest effects on classifier accuracy. This experiment introduces basic defender motions to illustrate the general classification effects of drastically different movement patterns. Defender motion (DM) was varied across five distinct types: \textit{Star} (basis for the original paper), \textit{Semi} (semi-circle), \textit{PerpL} (perpendicular left), \textit{PerpR} (perpendicular right), and \textit{Straight}. These motions were chosen to represent the primary vector components relative to the threat axis of 45$^\circ$, including normal (perpendicular left and right), tangential (straight), as well as varying allowable defender heading spread limits centered along the threat axis (e.g., straight = $45^\circ\pm 0^\circ$, star = $45^\circ\pm 45^\circ$, and semi = $45^\circ\pm 90^\circ$). Our aim was to investigate the effects of motions that are markedly different from one another. For visualization, Figure \ref{fig:dm_each} highlights the unique adversary responses (e.g. NN inputs) at time step 20 caused by the different DM, even when using identical engagement initialization.

\subsubsection{VOI \#3 - Measurement Noise}
Measurement noise, inherent in the radar measurements whether due to atmospheric conditions or adversarial jamming, is a highly likely occurrence in real-world engagements. To account for this, the noise level is varied from 0 to 50 in increments of 1. This VOI is unique in demonstrating the applicability of this dataset enrichment technique not only for controllable DOF but also for uncontrollable factors, such as expected environmental variations. The mathematical approach for introducing noise involved adding zero-mean Gaussian noise to the measurements, as described in our original work.

\subsubsection{Combining Datasets Considerations}
\paragraph{Exponential Increase in Dimensionality}
A significant challenge encountered when enriching a dataset by combining sub-datasets is the curse of dimensionality. Each dataset variable introduced adds an additional dimension to the dataset space. As more variables are considered, the number of possible combinations increases exponentially, leading to an explosion in the number of sub-datasets. If every possible point across all variables were to be combined into a single dataset, the resultant dataset would be extremely large, posing significant computational challenges. To manage this complexity, each VOI dimension is considered independently, with all other variables held at reference values. This approach allows us to systematically evaluate the impact of each VOI on classification performance.
\paragraph{Consistent Scaling}
Another significant consideration when joining sub-datasets into a ``combined" dataset, especially when training on one dataset and then testing on different datasets, is consistent input data scaling. Without common scaling, disparities in performance might arise not due to the NN's ability to generalize across different conditions, but because of input data scaling inconsistencies. To ensure uniformity, for each experiment identical scaling was applied to all datasets. This method reduces the risk of the NN learning to differentiate between datasets based on scaling differences rather than meaningful variations in the data.

\subsection{Neural Network Parameters}

A single baseline NN architecture, input length, and output type was used for all experiments. From the five top NNs developed in our original work, the Convolutional NN (CNN) was selected for its small memory size and shorter training time. Similarly, NN inputs of length 20 time steps were used, as longer time length inputs increased computation time for marginal accuracy improvements. However, a few CNN model variations were used, including a slightly deeper CNN for motion and noise experiments, increasing NN capacity to help with these more complex datasets, as well as a longer time input for the noise experiment. Regarding NN output, multiclass was chosen over multilabel due to its larger capacity for improvement, helping to identify and amplify the impact of different variables. Furthermore, for the noise experiment, a 50 time step input and a multihead output architecture was trained, however only the multiclass output was used for evaluation. These variations offered fair comparison with the noise results from our previous paper. Table \ref{tab:cnn_parameters} summarizes the different CNN model variations used for each experiment.

\begin{table}[h!]
    \renewcommand{\arraystretch}{1.3}
    \caption{CNN Parameters vs. Experiment}
    \label{tab:cnn_parameters}
    \centering
    \begin{tabularx}{\columnwidth}{l!{\vrule width 1.5pt}>{\centering\arraybackslash}X|>{\centering\arraybackslash}X|>{\centering\arraybackslash}X}
        \Xhline{3\arrayrulewidth} 
        \textbf{CNN Parameter} & \textbf{Defender Number} & \textbf{Defender Motion} & \textbf{Measurement Noise} \\
        \Xhline{3\arrayrulewidth} 
        Input Time & 20 & 20 & 50 \\
        \hline
        Layer Filters & 32 & 64,64,64 & 64,64,64 \\
        \hline
        Kernel Size & 7 & 7,5,3 & 7,5,3 \\
        \hline
        Pool Size & 5 & 3 & 3 \\
        \hline
        Dropout & 0.1 & 0.4  & 0.4 \\
        \Xhline{2\arrayrulewidth} 
    \end{tabularx}
\end{table}

\subsection{Optimizing Defender Trajectories for a Trained NN}

\begin{figure}[h]
    \centering
    \includegraphics[width=\linewidth]{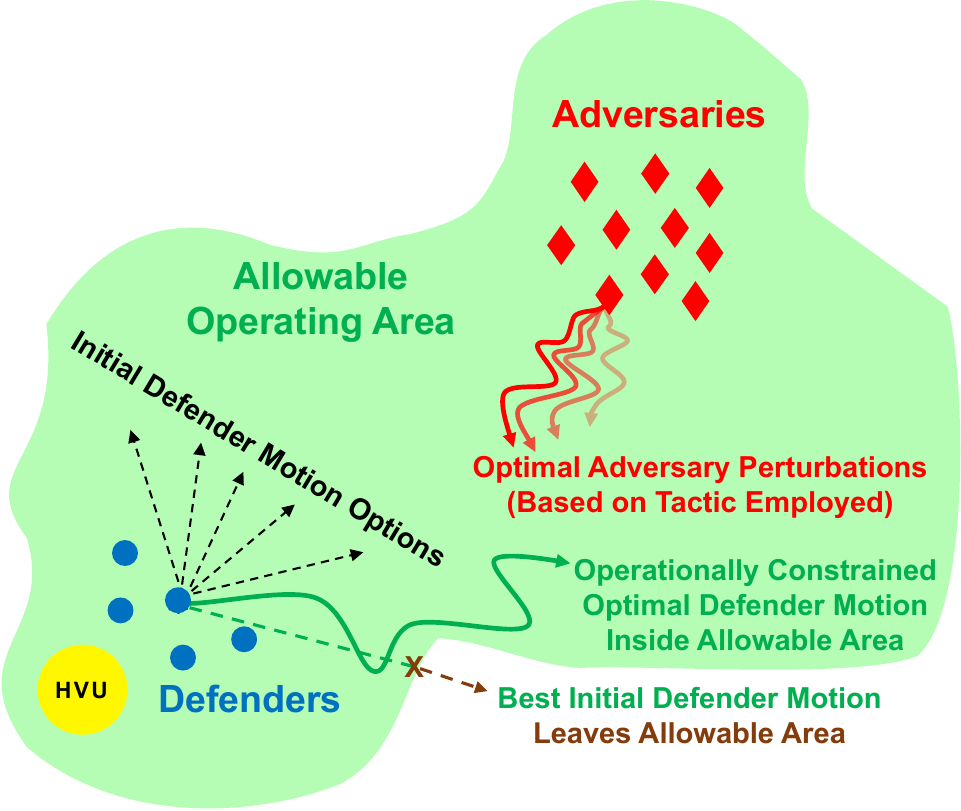}
    \caption{Overview of the trajectory optimization process for one defender resulting in an optimized response from the adversary regardless of tactic employed.}
    \label{fig:odm_scenario}
\end{figure}

After training a robust NN on diverse defender motions, enabling the NN to handle a wider variety of adversarial trajectories, the next key objective of this work is to develop a framework for deploying the NN in an operational environment. Given a trained NN we seek to generate defender trajectories that will maximize performance of the NN while satisfying operational constraints on the defender motion, for example airspace restrictions. Figure \ref{fig:odm_scenario} shows an example scenario, where defenders surrounding a high value unit (HVU) engage in maneuvers that optimize adversary tactic classification by the NN while respecting operational constraints, namely remaining inside the allowable operating area.

In this paper we propose to cast the problem of determining required defender motion as an optimal control problem. Let \( N_D \) be the number of defenders, and suppose \( P_j \in \mathbb{R}^{N_t \times 2}, j = 1:N_D \) represents the 2D position history (i.e. trajectory) of defender \( j \) during the discretized observation window \( t \in [t_0, t_f] \), where $N_t$  is the length of \( [t_0, t_f] \). Furthermore, let \( P_D = \left[P_1 \, \cdots \, P_{N_D}\right] \in \mathbb{R}^{N_t \times 2N_D} \) represent the matrix of all defender trajectories. Similarly, let \( N_A \) be the number of adversaries, and \( {P_A}_k(P_D) \in \mathbb{R}^{N_t \times 2N_A} \) represent the matrix of all adversary trajectories in response to defender trajectories \( P_D \) when a tactic \( k \) is employed by each adversary. Finally, let \( \mathcal{P}_k = \text{NN}({P_A}_k)\) represent the neural network’s predicted probability for the tactic \( k \), also referred to as the \textit{true prediction}, when the input to neural network is given by adversary trajectories \( {P_A}_k \).

For \( n \) possible adversary tactics, the NN output is a probability vector of length \( n \), with all elements summing to 100\%. For example, if the NN output is [94 1 3 2] for four possible  tactics used by the adversaries, and tactic one was in fact used, the true prediction is \( \mathcal{P}_1 = \text{NN}({P_A}_1) = 94\% \). We use the \textit{Sum of True Predictions} (STP) to define the performance of the NN to be maximized. Thus, STP is computed by summing  true predictions for each  adversary tactic. Note, the maximum value of STP is \( \text{Maximum STP} = n*100 \) (i.e. Maximum STP = 400 given four possible tactics).

An alternate STP formulation is stacking the NN output vectors for all possible adversary tactics (responses) and taking the trace of the resulting matrix. Let \( \mathcal{P}_{\text{stack}} \) be an \( n \times n \) matrix where each row represents the neural network’s output (i.e. sums to 100\%) when a  tactic \( k \) is used. Each entry \( \mathcal{P}_{kl} \) in \(\mathcal{P}_{\text{stack}} \) represents the predicted probability of the tactic \( l \) when tactic \( k \) is the input.

\[
\mathcal{P}_{\text{stack}} = 
\begin{bmatrix}
\mathcal{P}_{11} & \mathcal{P}_{12} & \dots & \mathcal{P}_{1n} \\
\mathcal{P}_{21} & \mathcal{P}_{22} & \dots & \mathcal{P}_{2n} \\
\vdots & \vdots & \ddots & \vdots \\
\mathcal{P}_{n1} & \mathcal{P}_{n2} & \dots & \mathcal{P}_{nn} \\
\end{bmatrix} =
\begin{bmatrix}
\mathcal{P}_{1} & \mathcal{P}_{12} & \dots & \mathcal{P}_{1n} \\
\mathcal{P}_{21} & \mathcal{P}_{2} & \dots & \mathcal{P}_{2n} \\
\vdots & \vdots & \ddots & \vdots \\
\mathcal{P}_{n1} & \mathcal{P}_{n2} & \dots & \mathcal{P}_{n} \\
\end{bmatrix}
\]

The matrix \( \mathcal{P}_{\text{stack}} \) captures the probability distribution across all tactics for each possible NN input (i.e. each possible tactic used). The STP, which is a function of all possible adversary trajectories \( {P_A}_k \) determined by a single set of defender trajectories \( P_D \), can now be expressed by taking the trace of \( \mathcal{P}_{\text{stack}} \).

\[
\text{STP}(P_D) = \operatorname{tr}(\mathcal{P}_{\text{stack}}) = \sum_{k=1}^{n} \mathcal{P}_{kk} =\sum_{k=1}^{n} \mathcal{P}_k
\]

Next, we assume that the dynamics of a defender \( j \) can be represented using \eqref{eqn: def vel} and \eqref{eqn: def pos}.

\begin{equation}
\label{eqn: def vel}
V_j(t+1) = V_j(t) + \Delta t \cdot A_j(t)
\end{equation}
\begin{equation}
\label{eqn: def pos}
P_j(t+1) = P_j(t) + \Delta t \cdot V_j(t)
\end{equation}

where:
\begin{description}
    \item[] \(A_j \) is acceleration  vector,
    \item[] \( \Delta t \) is sampling period.
\end{description}
\vspace{\baselineskip}

Since \eqref{eqn: def vel} and \eqref{eqn: def pos} represent a double integrator one can easily compute velocity and acceleration of each defender using its position.

\[
V_D(t) = \frac{P_D(t+1) - P_D(t)}{\Delta t}, \quad t = 0, \ldots, N_t-2
\]
\[
A_D(t) = \frac{V_D(t+1) - V_D(t)}{\Delta t}, \quad t = 0, \ldots, N_t-3
\]

Now, an optimal control problem to determine optimal defender trajectories \( {P_D}^* \) that maximize the NN performance metric (\( \text{STP} \)), while satisfying defender dynamics, operational area, and collision avoidance constraints can be written. Note that the negative of STP is used as the cost function in the optimal control problem formulation. 

\[
\min_{P_D} \{J = -\text{STP}(P_D)\}
\]

subject to:
\[
\begin{alignedat}{2}
P_D(t_0) &= {P_D}^0 &\quad &\text{Initial positions fixed} \\
P_D(t) &\in \mathcal{X} &\quad &\text{Operational constraints} \\
V_{\min} - \|V_D(t)\| &\leq 0 &\quad &\text{Min velocity constraint} \\
\|V_D(t)\| - V_{\max} &\leq 0 &\quad &\text{Max velocity constraint} \\
\|A_D(t)\| - A_{\max} &\leq 0 &\quad &\text{Max acceleration constraint} \\
\| P_i(t) - P_j(t)\| &\ge d_{\min}& \\ \forall \;\text{defenders} \;i,j, &\;\;i \ne j 
&\quad &\text{Collision avoidance constraint}
\end{alignedat}
\]
\vspace{\baselineskip}

where:

\begin{description}
    \item[] \( t \in [t_0, t_f] \) is time observation window,
    \item[] \( P_D \) are defender trajectories,
    \item[] \( {P_D}^0 \) is the vector of initial defender postions,
    \item[] \( \mathcal{X} \) represents operational constraints
    \item[] \( V_D \) are defender velocities,
    \item[] \( V_{\min} \) and \( V_{\max} \) are the velocity limits,
    \item[] \( A_D \) are defender accelerations,
    \item[] \( A_{\max} \) is the acceleration limit,
    \item[] \( d_{\min} \) is the minimum separation distance,    
    \item[] \( \|\cdot{}\| \) represents the vector 2 norm.
\end{description}
\vspace{\baselineskip}

Standard nonlinear programming (NLP) techniques were employed to obtain a numerical solution to the optimal control problem \cite{nocedal_numerical_2006}. Algorithm 1 was used to generate initial guess for the optimization.

\subsection{Minimum Number of Defenders Required}

In an operational environment quite often the number of defenders available for any mission is limited. Therefore, a natural question that an operational planner may ask is what is the minimum number of defenders required to successfully determine the tactics used by an adversarial swarm. The numerical solution to the optimal control problem presented in the previous section can easily be used to obtain an answer to this question. The details are shown in Section \ref{sec:results min num def}.

\section{Experimental Setup}
\label{sec:exp setup}

The software and hardware used were consistent with those described in the original paper \cite{peltier_swarm_2024}. All NN training and evaluation were conducted using Python and TensorFlow. The MATLAB function \texttt{fmincon} was used to perform numerical optimization \cite{mathworks_inc_fmincon_2024}. For reproducibility, hyperparameters for NN training and optimizer options are available in the public code repository\footnote{https://github.com/DWPeltier3/Swarm-NN-TSC}.

\subsection{Generate Enriched Datasets}

To investigate the impact of defender number ($N_D$) and defender motion (DM), the existing simulation environment was updated to incorporate these new variables of interest. Defender number was straightforward to adjust, and defender motions were coded as described in Algorithm \ref{alg:defender_motion} to reflect the basic motions shown in Figure \ref{fig:dm_each}. Measurement noise datasets were created by adding varying amounts of zero-mean Gaussian noise to the original dataset.

\begin{algorithm}
\algofontsize 
\caption{Defender Motion Algorithm}
\label{alg:defender_motion}
\begin{algorithmic}[1]
\Require Defender number, motion type, and velocity constraints
\Ensure Defender headings and velocities
\State Initialize defender velocity spread as the difference between maximum and minimum velocities
\State Randomly assign each defender a velocity magnitude within the velocity spread 
\If{motion type is "Star"}
    \State randomly assign each defender a heading between 0 and 90 degrees (between East and North)
\ElsIf{motion type is "Semi-circle"}
    \State randomly assign each defender a heading between -45 and 135 degrees (between South-East and North-West)
\ElsIf{motion type is "Straight"}
    \State Set all defender headings to 45 degrees (North-East)
\ElsIf{motion type is "Perpendicular Left"}
    \State Set all defender headings to 135 degrees (North-West)
\ElsIf{motion type is "Perpendicular Right"}
    \State Set all defender headings to -45 degrees (South-East)
\EndIf
\State Compute each defender x and y components of velocity based on individual heading and velocity magnitude
\end{algorithmic}
\end{algorithm}

The dataset generation process used the methodology summarized in the original paper, simulating engagements to generate NN inputs consisting of adversary positions and velocities ordered by time, along with corresponding tactic labels. For each VOI independently ($N_D$, DM, and Noise), separate sub-datasets were generated at each point along the VOI dimension (see Section \ref{sec:dataset enrichment} for VOI space descriptions), with all other variables held at reference. All associated sub-datasets belonging to a particular VOI were joined into a ``Combined” dataset. In order to combine datasets, the minimum time across all associated VOI datasets was determined and used to truncate all instances equally, ensuring uniform dataset dimensions.

Another thrust in evaluating defender motion optimization was exploring the effect of different trained NNs on the optimization results. Therefore, in addition to the NN trained on the ``Combined DM" dataset, a NN was trained on an enriched dataset with an order of magnitude more training instances. In our original work, limiting the number of engagement instances was a self-imposed constraint to facilitate NN performance comparisons. However, for real-world deployment, a NN would use as many engagement instances as needed to achieve the desired performance level. As expected, this larger dataset produced a NN with superior classification performance, and therefore was labeled ``Combined DM+".

Lastly, to explore the minimum number of optimized defenders required, a NN robust enough to handle variations in both defender motion and defender number was needed. Therefore, two VOI dimensions were merged (i.e., five basic motions and $N_D=[1:10]$), resulting in the ``Combined $N_D$ \& DM" dataset, which combines 50 sub-datasets.

Tables \ref{tab:dataset_summary_1} and \ref{tab:dataset_summary_2} summarize and compare the original dataset with the enriched ``Combined" VOI datasets, with parameters in \textbf{bold} indicating deviations from the reference values.
\begin{table}[h!]
    \renewcommand{\arraystretch}{1.3}
    \caption{Datasets Summary (Robustness Experiments)}
    \label{tab:dataset_summary_1}
    \centering
    \begin{tabularx}{\columnwidth}{l!{\vrule width 1.5pt}>{\centering\arraybackslash}X|>{\centering\arraybackslash}X|>{\centering\arraybackslash}X|>{\centering\arraybackslash}X}
        \Xhline{3\arrayrulewidth} 
        \textbf{Parameter} & \textbf{Original Dataset} & \textbf{Comb $\bm{N}_{\bm{D}}$} & \textbf{Comb DM} & \textbf{Comb Noise} \\
        \Xhline{3\arrayrulewidth} 
        Noise Level & 0 & 0 & 0 & \textbf{0 to 50} \\
        \hline
        Motions & star & star & \textbf{all 5} & star \\
        \hline
        $N_D$ & 10 & \textbf{1 to 15} & 10 & 10 \\
        \hline
        $N_A$ & 10 & 10 & 10 & 10 \\
        \Xhline{3\arrayrulewidth} 
        Max Time Steps & 58 & 45 & 50 & 58 \\
        \hline
        Total Instances & 4,800 & 72,000 & 24,000 & 244,800 \\
        \Xhline{3\arrayrulewidth} 
        Training (60\%) & 2,880 & 43,200 & 14,400 & 146,880 \\
        \hline
        Validation (15\%) & 720 & 10,800 & 3,600 & 36,720 \\
        \hline
        Test (25\%) & 1,200 & 18,000 & 6,000 & 61,200 \\
        \hline
    \end{tabularx}
\end{table}

\begin{table}[h!]
    \renewcommand{\arraystretch}{1.3}
    \caption{Datasets Summary (Optimization Experiments)}
    \label{tab:dataset_summary_2}
    \centering
    \begin{tabularx}{\columnwidth}{l!{\vrule width 1.5pt}>{\centering\arraybackslash}X|>{\centering\arraybackslash}X|>{\centering\arraybackslash}X|>{\centering\arraybackslash}X}
        \Xhline{3\arrayrulewidth} 
        \textbf{Parameter} & \textbf{Original Dataset} & \textbf{Comb DM} & \textbf{Comb DM+} & \textbf{Comb $\bm{N}_{\bm{D}}$\&DM} \\
        \Xhline{3\arrayrulewidth} 
        Noise Level & 0 & 0 & 0 & 0 \\
        \hline
        Motions & star & \textbf{all 5} & \textbf{all 5} & \textbf{all 5} \\
        \hline
        $N_D$ & 10 & 10 & 10 & \textbf{1 to 10} \\
        \hline
        $N_A$ & 10 & 10 & 10 & 10 \\
        \Xhline{3\arrayrulewidth} 
        Max Time Steps & 58 & 50 & 50 & 45 \\
        \hline
        Total Instances & 4,800 & 24,000 & 200,000 & 240,000 \\
        \Xhline{3\arrayrulewidth} 
        Training (60\%) & 2,880 & 14,400 & 120,000 & 140,000 \\
        \hline
        Validation (15\%) & 720 & 3,600 & 30,000 & 36,000 \\
        \hline
        Test (25\%) & 1,200 & 6,000 & 50,000 & 60,000 \\
        \hline
    \end{tabularx}
\end{table}

\subsection{Neural Network Training, Evaluation, and Conversion}

To ensure consistent NN input scaling, all datasets were scaled during preprocessing using the mean and variance derived from the combined VOI dataset. As such, NN training and inference functions were modified to apply this common scaling across all sub-datasets.

For the VOI robustness experiments ($N_D$, DM, Noise), individual NNs were trained on each distinct VOI sub-dataset (e.g., $N_D=1$, $N_D=2$, etc.), as well as on the combined dataset comprising all VOI sub-datasets (i.e., Combined $N_D$). The NN trained on the combined dataset was denoted as the \textit{robust} NN for that particular VOI. Subsequently, all NNs associated with a particular VOI experiment were evaluated by performing inference on each corresponding VOI sub-dataset test set, allowing for direct performance comparisons across the different NNs.

For the optimization experiments, pretrained TensorFlow neural network models were imported into MATLAB using \texttt{importNetworkFromTensorFlow} \cite{mathworks_inc_importnetworkfromtensorflow_2024}. It is important to note that this conversion tool did not support transformer architectures, which bolstered the decision to use a CNN model. The defender motion optimization experiment required two robust NNs, ``Comb DM" and ``Comb DM+", while the minimum number of optimized defenders experiment required only one robust NN, ``Comb $N_D$ \& DM". In addition to converting the NN for use in MATLAB, the mean and variance matrices associated with each combined dataset were also converted to ensure proper input scaling during NN inference in MATLAB.

\subsection{Optimize Defender Motion}

The first goal of the optimization experiment was proof of concept. Various combinations of different engagement initial conditions, allowable operating area constraints, and initial defender motions were optimized to demonstrate the flexibility of our optimal defender motion framework. Next, optimized variations for a single engagement are compared, including all five basic defender motions with identical allowable operating areas, and the influence of different allowable area constraints and defender initial trajectory combinations. Finally, optimizations using the two different motion-robust NNs were generated for all five basic defender motions under the same allowable operating area. Overall, to ensure that the engagements considered during experimentation were not part of the datasets used to train a particular NN, random generator seeds above 1,200 or 10,000 were used, depending on the NN.

\subsection{Finding Minimum Number of Optimal Defenders}

To effectively demonstrate that the number of defenders could be minimized using optimal motion, the optimal defender motion framework was applied to each basic defender motion while varying the number of defenders from one to ten. Additionally, varying whether or not defenders accelerated to their maximum velocity during the initial trajectory generation affected both the initial and optimized STP, providing an additional variable to help improve optimization.

\section{Results \& Analysis}
\label{sec:results}

Our first goal was to demonstrate how a robust NN could improve performance during uncertain operational conditions, particularly focusing on the number of defenders, their motion, and measurement noise. Our second goal was to create a NN input optimization framework that provides planners with a tool to automate the use of the aforementioned robust classifier.

\subsection{Defender Number Robust NN}

We initially assumed that a NN trained with a higher number of defenders would perform best during inference, as more defenders would better motivate each individual adversary to maneuver. However, the results revealed several interesting patterns.

\begin{figure}[h]
    \centering
    \includegraphics[width=\linewidth]{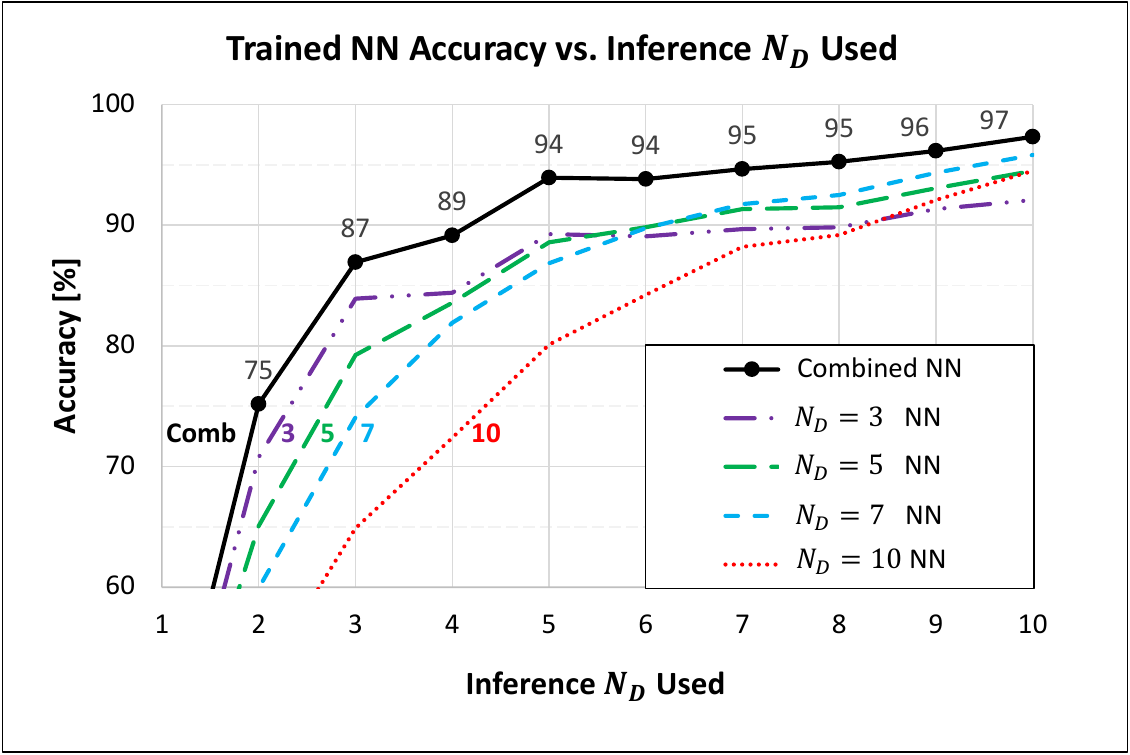}
    \caption{Defender Number Experiment: comparison of 5 NN; four NN trained with fixed defender number ($N_D$), and one NN trained using ``Combined" dataset ($N_D=[1:15]$).}
    \label{fig:dn_train_v_inference}
\end{figure}

Figure \ref{fig:dn_train_v_inference} illustrates that a NN trained with fewer defenders showed a more rapid increase in accuracy as the number of defenders used during inference increased. This phenomenon may be attributed to the fact that the relative improvement is more significant when adding defenders to a model trained on fewer defenders (e.g., going from 2 to 3 defenders is a 50\% increase in the number of defenders, compared to a 14\% increase when moving from 7 to 8 defenders). However, this rapid improvement came at the cost of a generally lower maximum achievable accuracy as the limit of 10 defenders was approached. For clarity, only a few fixed defender number ($N_D$) NNs are shown in Figure \ref{fig:dn_train_v_inference}, not all 15, to maintain readability.

\begin{figure}[h]
    \centering
    \includegraphics[width=\linewidth]{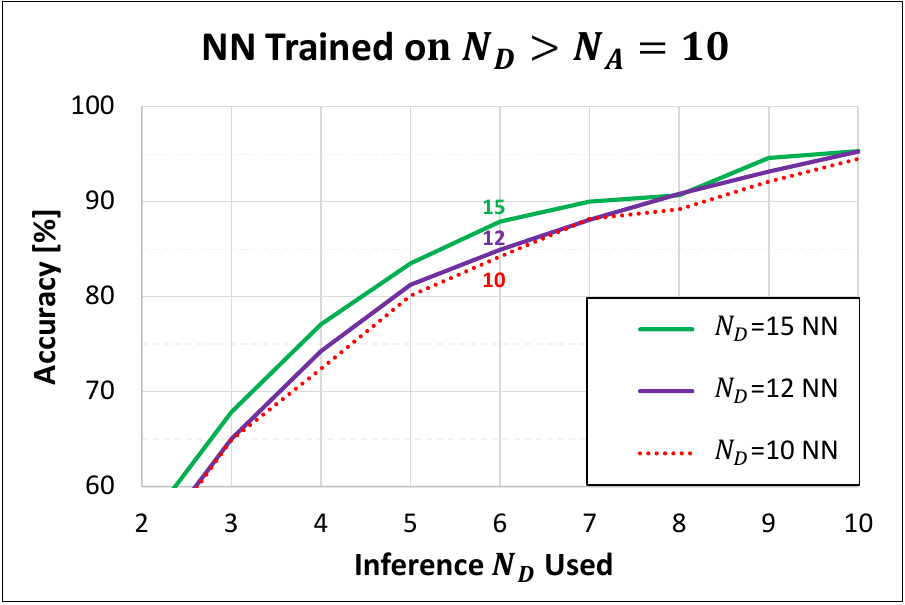}
    \caption{Accuracy improves slightly when a NN is trained with more defenders than adversaries.}
    \label{fig:dn_train_def_greater_adv}
\end{figure}

Unexpectedly, when a NN was trained with more defenders than adversaries (e.g., 15 defenders versus 10 adversaries), we observed a slight improvement in accuracy across every defender point, as shown in Figure \ref{fig:dn_train_def_greater_adv}. This outcome may be due to the extra defenders providing more options for adversary targeting at each time step (e.g., more targets in a fixed size space), which allowed for finer adjustments and quicker identification of adversary tactics.

\begin{figure}[h]
    \centering
    \includegraphics[width=\linewidth]{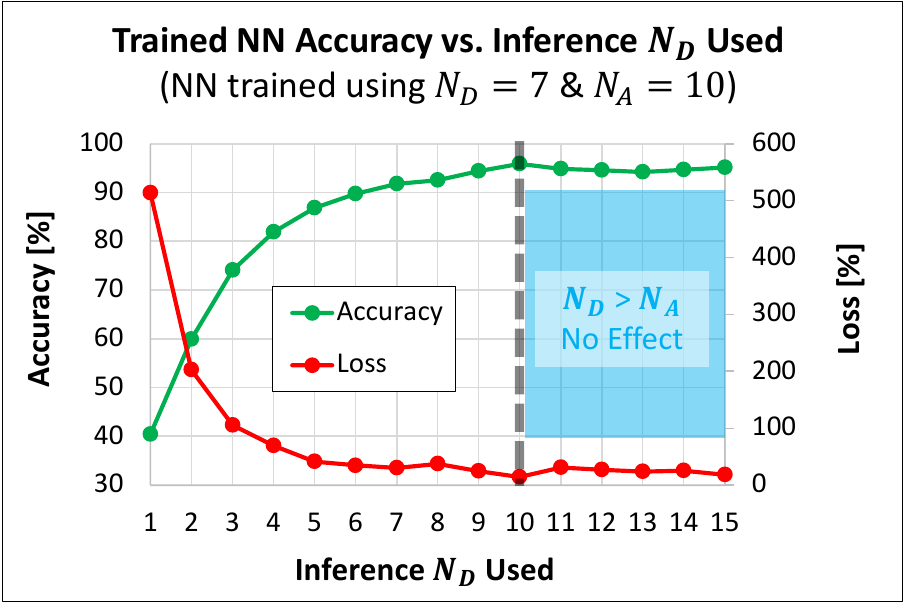}
    \caption{Accuracy plateaus when number of defenders exceeds number of adversaries during inference.}
    \label{fig:dn_inf_def_greater_adv}
\end{figure}

Similarly, we explored the effect of using more defenders than adversaries during inference. However, as illustrated in Figure \ref{fig:dn_inf_def_greater_adv} accuracy plateaued for all models when the number of defenders exceeded the number of adversaries during inference, which makes intuitive sense given that each adversary can only engage with a single targeted defender.

Overall, the most significant finding from the defender number VOI experiment was that the ``Combined $N_D$" robust NN consistently showed improved accuracy across all inference scenarios. Notably, it achieved relatively high accuracy (75-94\%) with substantially fewer defenders (e.g., 2-5 defenders, which represents an 80-50\% reduction from the original study's values), as shown in Figure \ref{fig:dn_train_v_inference}. This highlights the robust model's effectiveness under uncertainty, particularly when the number of available defenders is unknown, and demonstrates its potential for reducing resource expenditure. In other words, the combined model can minimize the number of assets required to effectively utilize the classifier.

\subsection{Defender Motion Robust NN}
\label{sec:results_DM}

\begin{figure}[h]
    \centering
    \includegraphics[width=\linewidth]{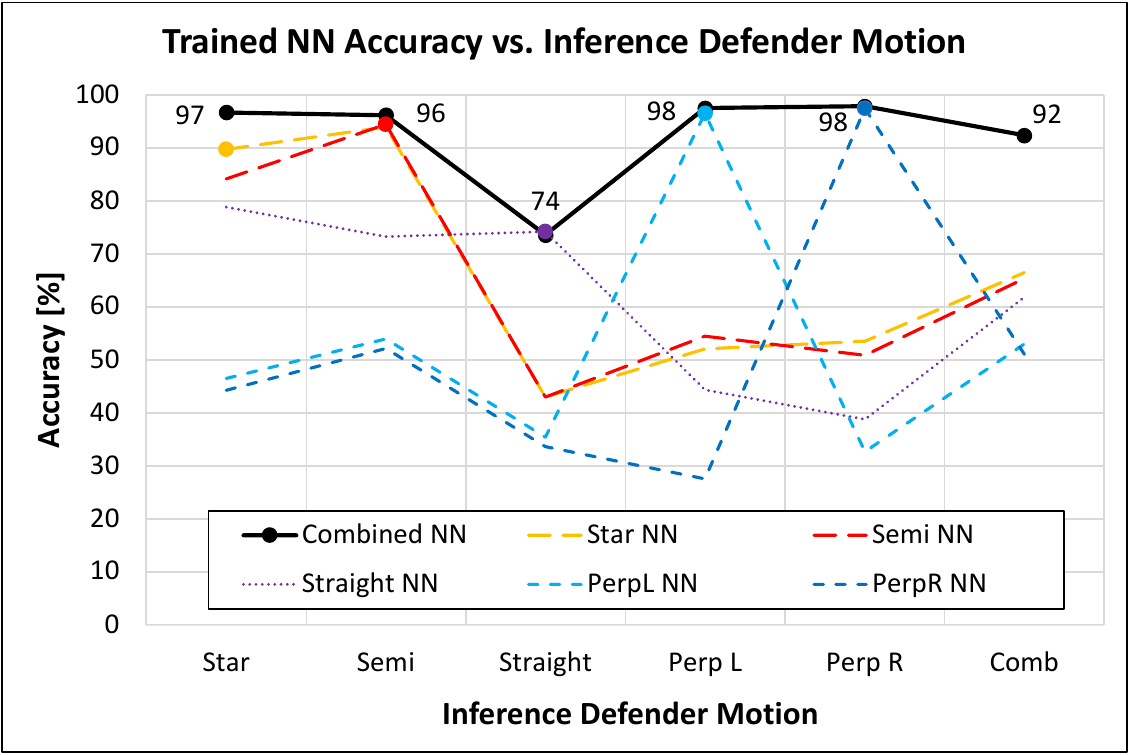}
    \caption{Defender Motion Experiment: comparison of 6 NN; five NN trained with one unique defender motion, and one ``Combined NN" trained using all five motions.}
    \label{fig:dm_train_v_inference}
\end{figure}

Our hypothesis was that defender motion would be a pivotal degree of freedom (DOF) and that we would identify a single optimal motion for training or inference. However, as seen in Figure \ref{fig:dm_train_v_inference}, a NN trained on a specific motion tends to develop a bias towards that particular motion. Specifically, the NN performs well on the motion it was trained on, as well as on similar motions, but struggles with dissimilar motions. For instance, the Star and Semi motions yielded comparable results, likely due to their similarity—the only difference being the allowable defender heading spread: $90^\circ$ for the star motion and $180^\circ$ for the semi motion. Conversely, Straight motion, where defenders fly directly at adversaries, almost universally resulted in the worst performance. This was likely due to the lack of significant maneuvering required from the adversaries, leading to poor activations, low accuracy, and higher loss.

An interesting outcome was that the ``Combined DM" robust NN achieved an accuracy greater than 90\% for all motions during inference, except for the straight motion. This result parallels the findings from the defender number experiment, suggesting that training on a diverse set of conditions enhances the NN's flexibility and performance. In the context of defender motion, this flexibility has several tactical implications. It allows improved classifier accuracy even when defender maneuvering constraints are imposed—such as near restricted airspace or waterways, or when deconfliction is needed to avoid collisions. Moreover, it enhances classifier accuracy when defender motion is not arbitrary but strategically significant, for instance, when specific motions are required for optimal tactical positioning to enhance unit survival or lethality. This flexibility enables defenders to classify adversary tactics quickly and incorporate this information into near-real-time counter-maneuver decisions.

\subsection{Measurement Noise Robust NN}

\begin{figure}[h]
    \centering
    \includegraphics[width=\linewidth]{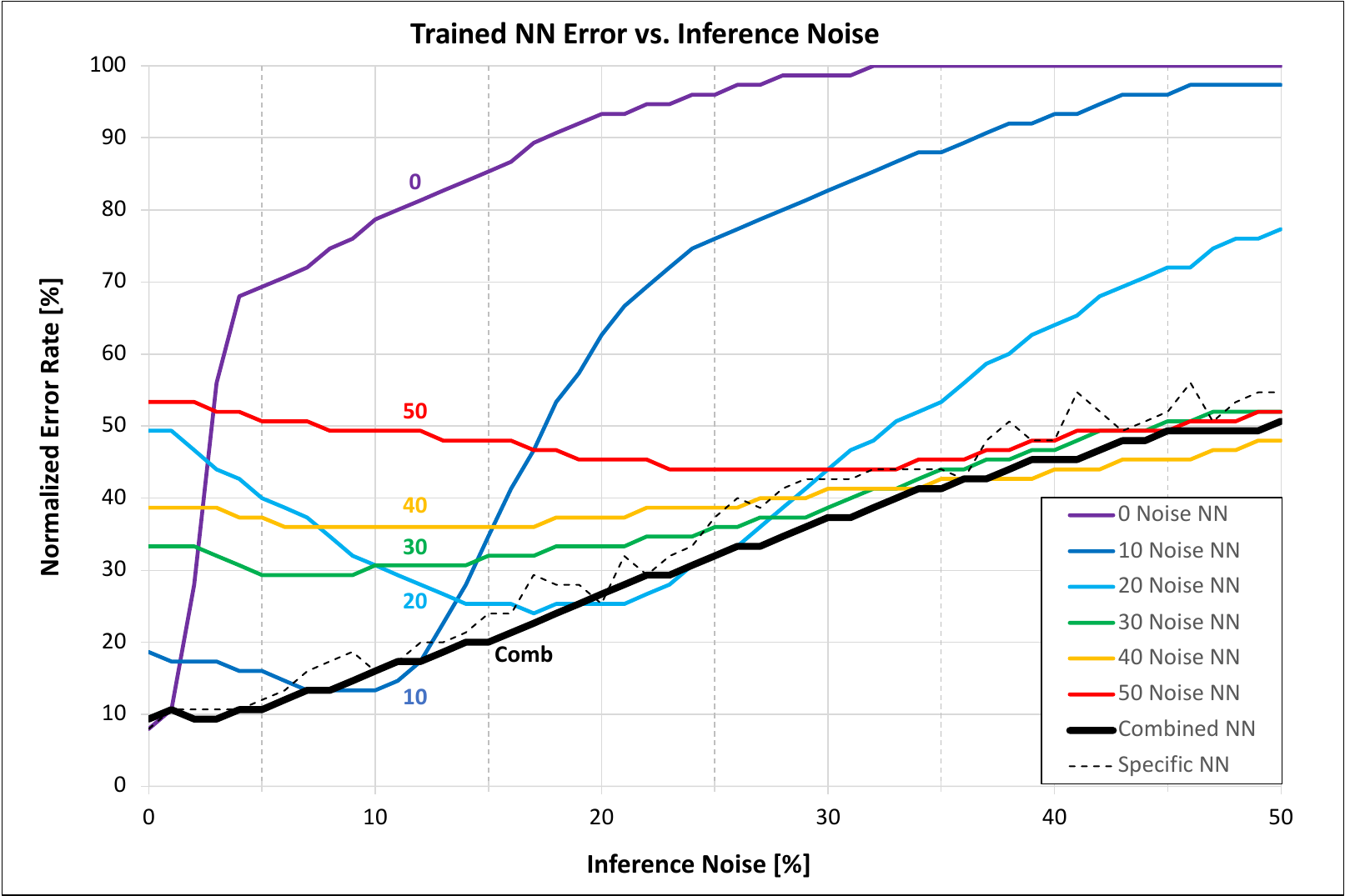}
    \caption{Comparison of NN performance during inference based on the noise level used for training. Solid lines represent a single NN, while the dotted line represents an ensemble of 51 NN.}
    \label{fig:noise_train_v_inference}
\end{figure}

In our previous work, we trained 51 NNs, each with a specific noise level, and found that the ensemble of these NNs exhibited graceful degradation as noise levels increased. In this current experiment, we trained NNs on \textbf{one} noise level and then evaluated their inference performance across various noise levels. Our findings, depicted in Figure \ref{fig:noise_train_v_inference}, indicate that as the training noise level increased, inference error tended to rise at lower noise levels but decrease at higher noise levels. This suggests that a model trained on higher noise levels might overcompensate on cleaner data while performing better under noisy conditions.

Notably, training on a ``combined" dataset that included multiple noise levels led to improved performance across most noise levels tested during inference. Once again, the approach of using a combined dataset enhanced the NN's robustness, enabling it to generalize better across varying conditions.

\subsection{Optimized Defender Motion}

\begin{figure}[h!]
    \centering
    \includegraphics[width=\linewidth]{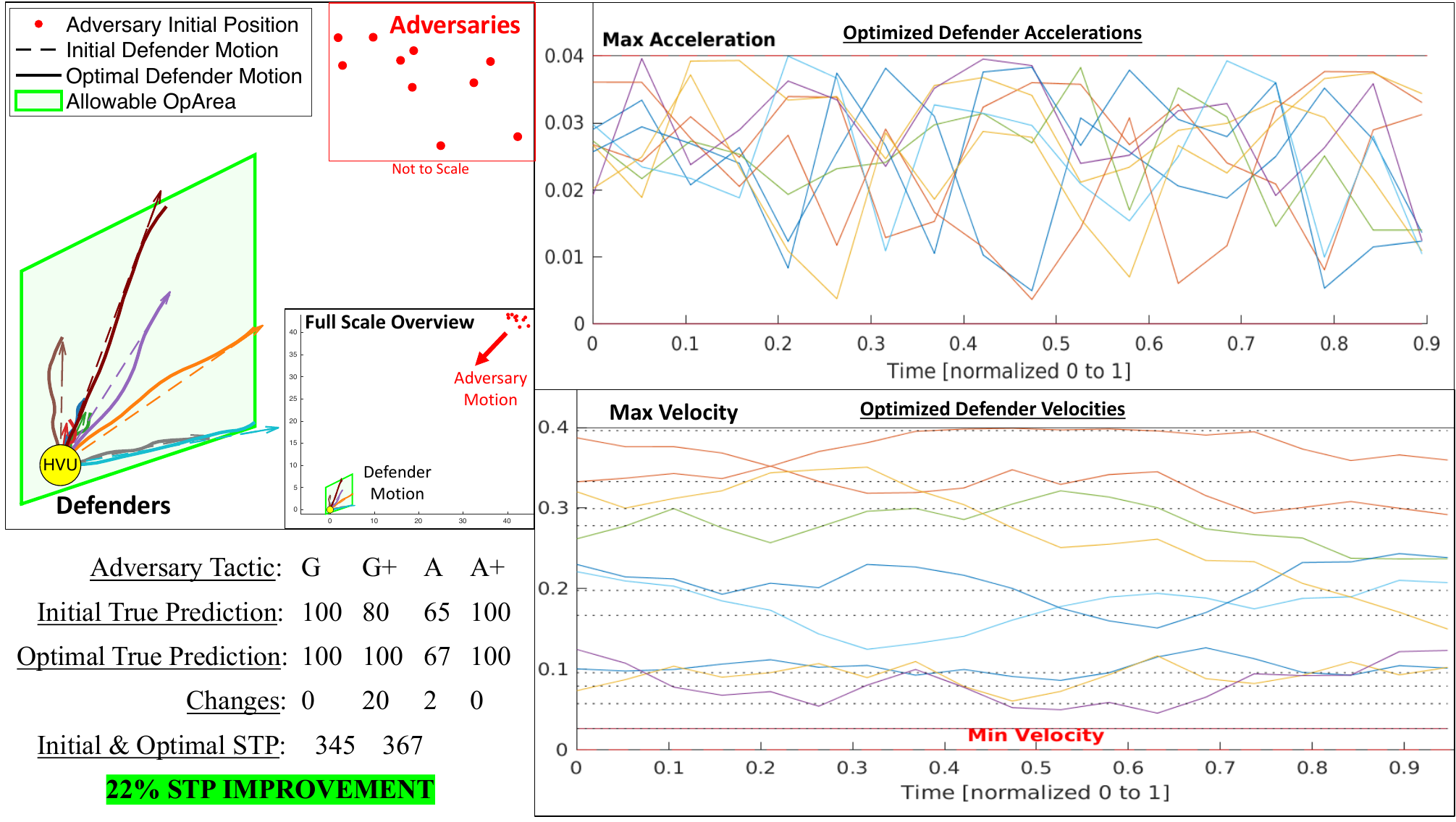}
    \caption{Defender trajectory optimization results for one engagement. Star defender trajectory yielded highest initial STP and optimization  increased an additional 22\%.}
    \label{fig:odm_example}
\end{figure}

This section presents the results of applying our optimal control framework to determine defender motion, effectively steering adversaries to maximize STP for a given trained NN classifier. First, to provide better visualization of the optimization process, Figure \ref{fig:odm_example} shows the outcome of optimizing defender trajectories for a single engagement scenario (seed 1202). This includes a full engagement overview, initial and optimized defender trajectories, constrained kinematic values for defender acceleration and velocity, and NN true predictions for each adversary tactic.

\begin{figure}[h!]
    \centering
    \includegraphics[width=\linewidth]{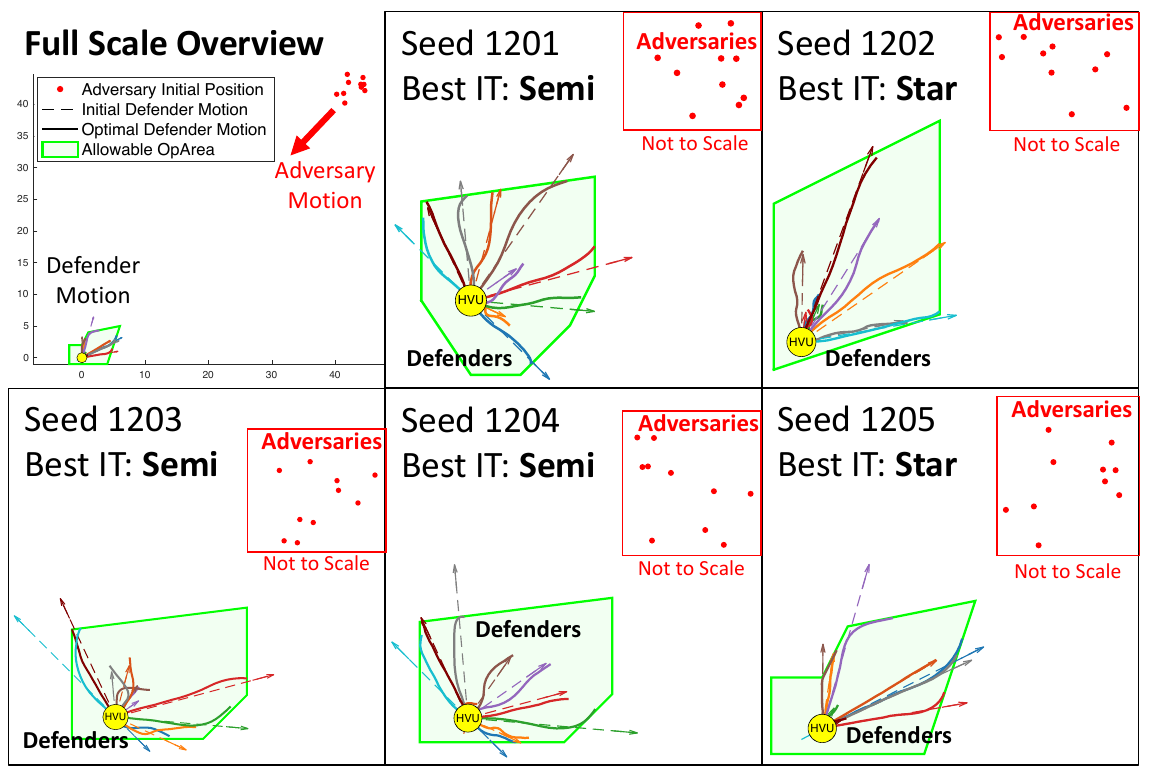}
    \caption{Best defender initial trajectory (IT) for various engagement initializations (i.e., random generator seeds).}
    \label{fig:odm_differentseeds}
\end{figure}

In fact, we aimed to explore how different engagement setups influenced the best initial guess for defender motion, and to determine which defender initial trajectory maximized the initial STP. By using different engagement setups (e.g., varying random seeds), we found that Semi or Star initial trajectory consistently yielded the highest initial STP, as illustrated in Figure \ref{fig:odm_differentseeds}. There are a few important points to note about this figure. First, the five seed plots are not to scale; adversaries are positioned closer to defenders to facilitate the visualization of the engagement setup. Second, the initial trajectory evaluation is not affected by the allowable area constraint or the optimized defender motions, which are shown merely as illustrative examples.

\begin{figure}[h!]
    \centering
    \includegraphics[width=\linewidth]{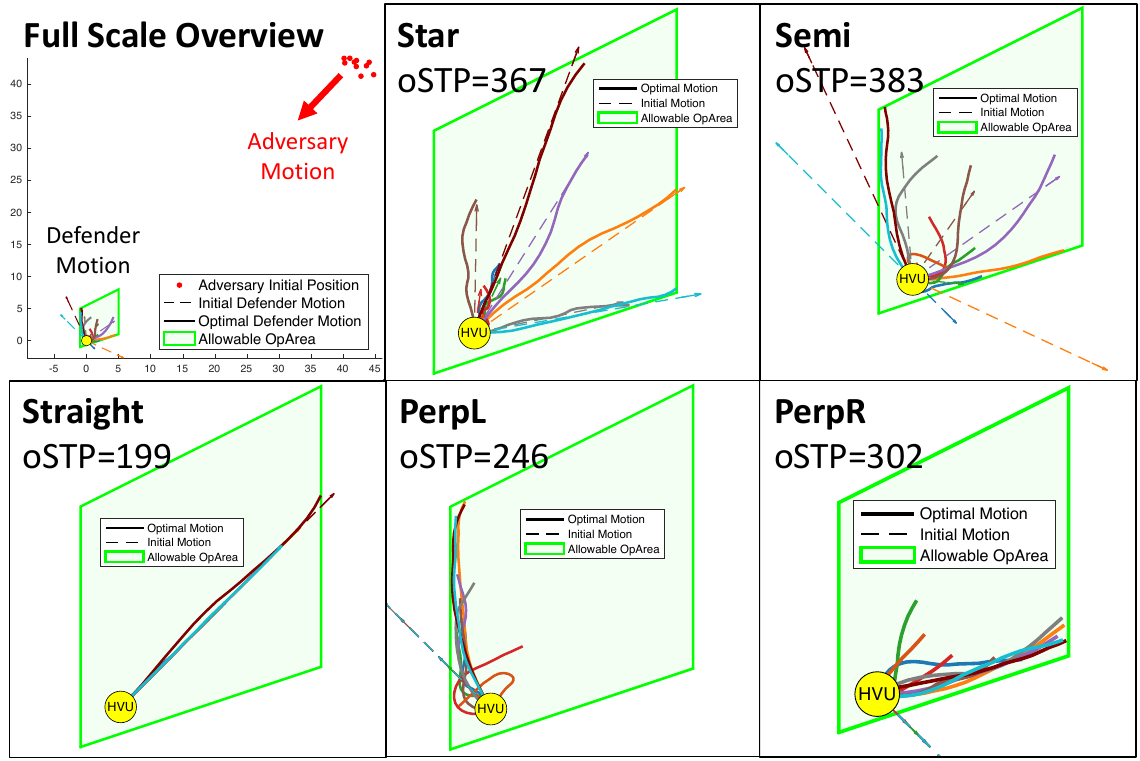}
    \caption{Optimized STP (oSTP) and defender trajectories using \textit{different} defender initial trajectories but same allowable area and engagement initial conditions.}
    \label{fig:odm_SamePolyDiffMotion}
\end{figure}

Continuing the analysis, we focused on one engagement scenario (seed 1202) and one operational area constraint to compare the effect of different defender initial trajectories on the optimization process. As shown in Figure \ref{fig:odm_SamePolyDiffMotion}, the Semi trajectory produced the highest optimized sum of true predictions (oSTP), even though the Star trajectory yielded the highest initial STP. This discrepancy suggests that initial and optimized STP do not always correlate, necessitating the optimization of all possible initial trajectories to guarantee finding the optimal defender trajectory. Importantly, while initial STP provides a measure for ranking different defender initial trajectories, it cannot guarantee resulting in the optimal defender trajectory due to potential violations of the allowable area constraint. Instead, optimized STP proved to be the more reliable metric.

\begin{figure}[h]
    \centering
    \includegraphics[width=\linewidth]{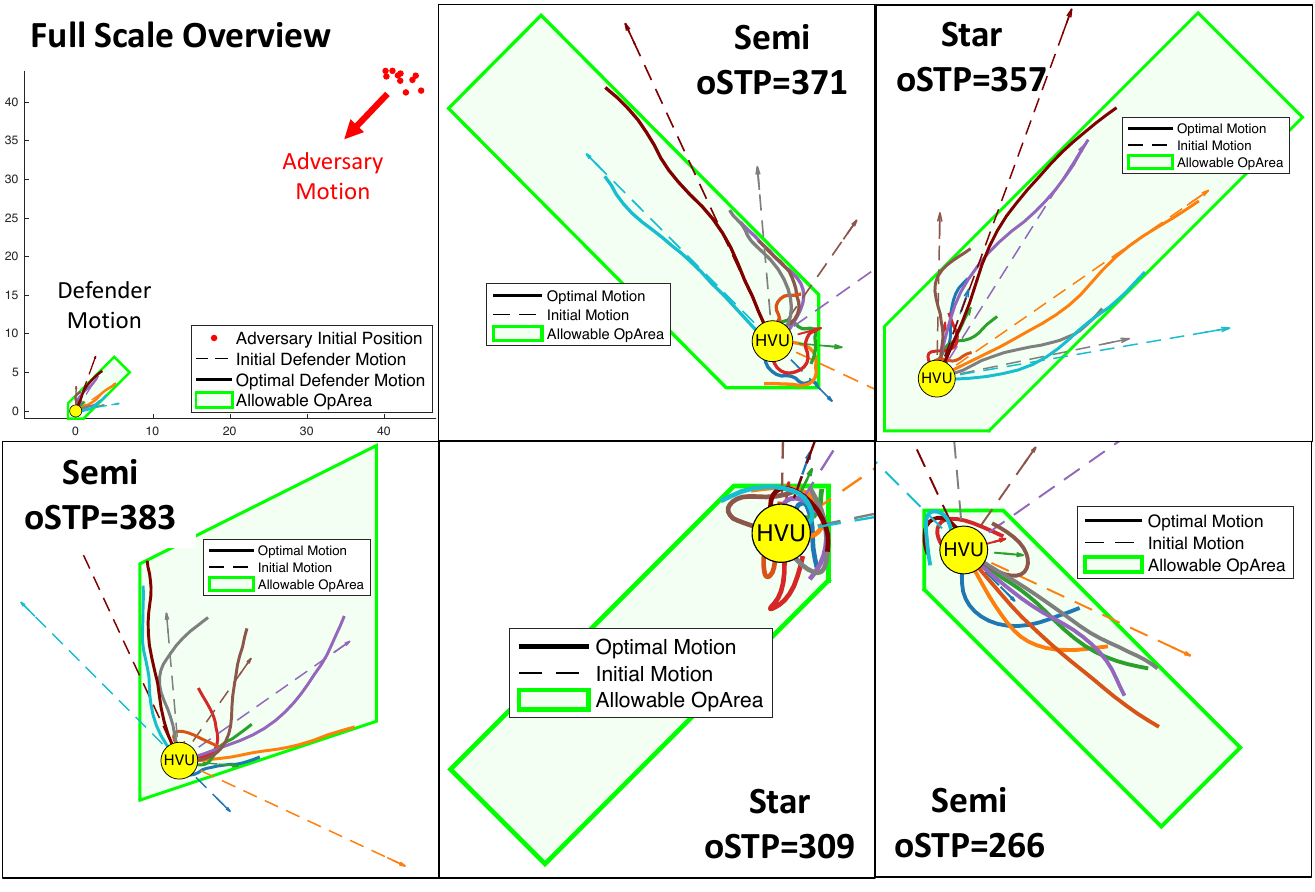}
    \caption{Optimized STP (oSTP) and defender trajectories for \textit{different} allowable area constraints.}
    \label{fig:odm_DiffPolyBestMotion}
\end{figure}

Our optimization framework was also applied to different allowable area constraints, demonstrating its adaptability to various operational boundaries. As shown in Figure \ref{fig:odm_DiffPolyBestMotion}, either Star or Semi initial defender trajectory consistently led to the best optimized motions for each scenario, reinforcing the observation made in Figure \ref{fig:odm_differentseeds} regarding the most effective initial defender trajectory. This finding suggests certain qualities that characterize the best initial defender motions. Specifically, both the Star and Semi configurations maximize the spread of defenders along the adversary threat axis of Northeast ($45^\circ$).

\begin{figure}[h]
    \centering
    \includegraphics[width=\linewidth]{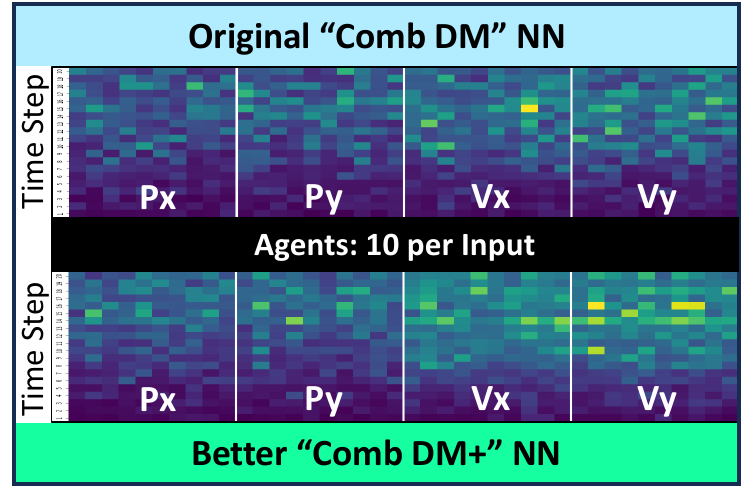}
    \caption{Comparing two different NN input to output gradient maps. The Better NN has more large gradients (indicated by lighter coloring) across both time and agents.}
    \label{fig:odm_NNsaliency}
\end{figure}

\begin{figure}[h]
    \centering
    \includegraphics[width=\linewidth]{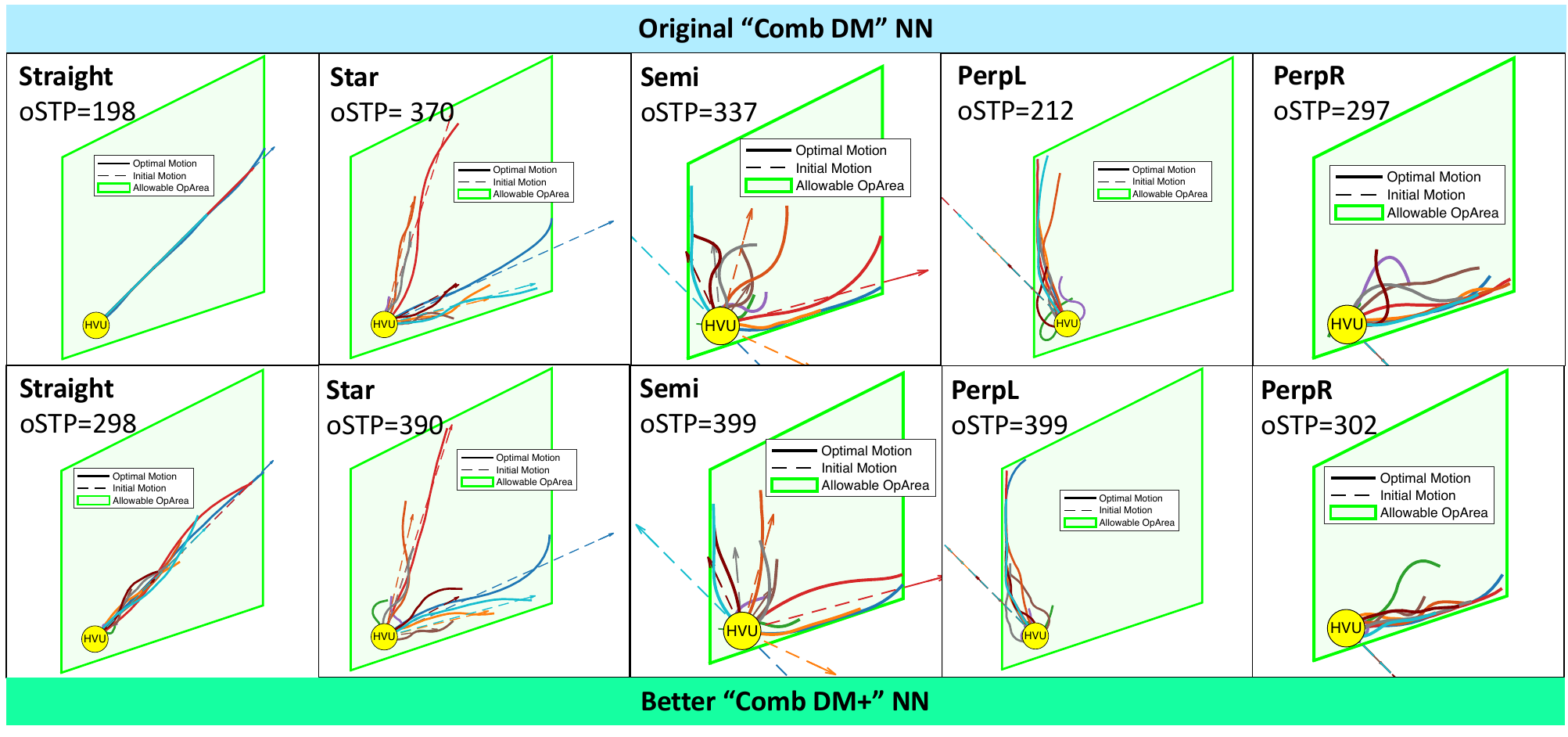}
    \caption{Optimization comparison using two different neural networks. The Better NN improves optimized STP (oSTP), regardless of initial motion; Maximum STP = 400).}
    \label{fig:odm_NNcompare}
\end{figure}

The final experiment compared optimization results for seed 10,002 using two different trained NNs. Figure \ref{fig:odm_NNsaliency} demonstrates that the ``Better NN" exhibited improved input-output gradients (i.e., saliency), with more pronounced gradients (indicated by lighter colors) spread across both agent and time dimensions. This enhanced saliency offers the motion optimizer more flexibility in improving STP, providing greater leverage from changing inputs across both agents and time steps. Consequently, as Figure \ref{fig:odm_NNcompare} shows, the optimal defender motion framework using the Better NN consistently improved STP outcomes, even achieving near-maximum optimized STP (oSTP = 399) using two distinct initial defender trajectories.

Overall, this analysis highlights the versatility and effectiveness of the NN performance optimization framework in adapting to operational constraints. The results demonstrate the ability of the framework to adapt to diverse engagement scenarios and consistently improve the sum of true predictions by optimizing defender motion. Moreover, the experiments indicate that the Star or Semi defender initial trajectories are generally the most effective, providing a useful heuristic to accelerate the optimization process in future engagements. Finally, enhancing the NN by increasing the number of generated training instances significantly improved the optimization results, allowing the process to yield near-perfect outcomes.

\subsection{Minimum Number of Optimized Defenders Required}
\label{sec:results min num def}

\begin{figure}[h]
    \centering
    \includegraphics[width=\linewidth]{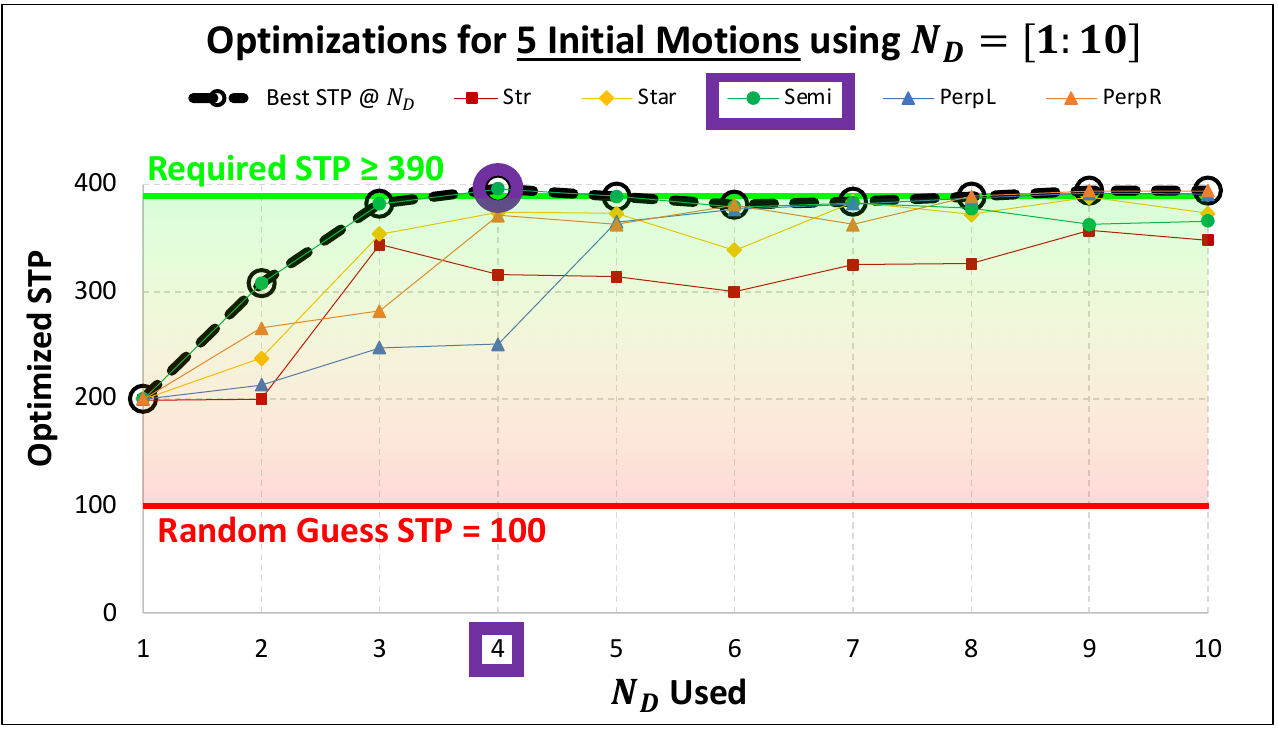}
    \caption{50 Optimizations for one engagement yields the minimum number of defenders needed for a required STP.}
    \label{fig:odm_dn}
\end{figure}

The goal of this section was to demonstrate that the number of defenders required could be minimized if optimized defender motion was employed. This was done by computing optimized defender trajectories for a varying number of defenders using a single engagement example. In addition, all five initial defender trajectories were used as initial guesses and the number of defenders was varied from 1 to 10. Figure \ref{fig:odm_dn} shows several interesting trends. First, using the straight initial motion generally yielded the worst optimized motion, which aligns with the findings and reasoning discussed previously. Second, the Star or Semi initial trajectories produced the best-performing optimized defender motion when five or fewer defenders were used. However, the perpendicular initial motions showed a steadily increasing optimized STP performance as defender number ($N_D$) increased, eventually surpassing the previously mentioned motions. Ultimately, by combining the highest STP at each $N_D$ used into a ``Best STP @ $N_D$" contour, an actionable performance estimate was obtained.

Operational planners and decision-makers can use the resulting ``Best STP @ $N_D$" estimate to make informed defensive decisions. For instance, if a required classifier performance minimum threshold is set, such as a Required STP $\geq 390$ to ensure a minimum NN prediction accuracy of 90\%, the Best STP plot can guide the selection of the optimal defender number and initial trajectory to meet these requirements. Additionally, if specific operational constraints are present, such as needing a particular defender motion, the Best STP plot can help determine the suitable minimum number of defenders for a given threshold of STP $\ge$ 390. Alternatively, the defender number and initial trajectory associated with the highest STP can be employed in a closed-loop context within a broader defender deployment operational tool to maximize effectiveness.

\begin{figure}[h]
    \centering
    \includegraphics[width=\linewidth]{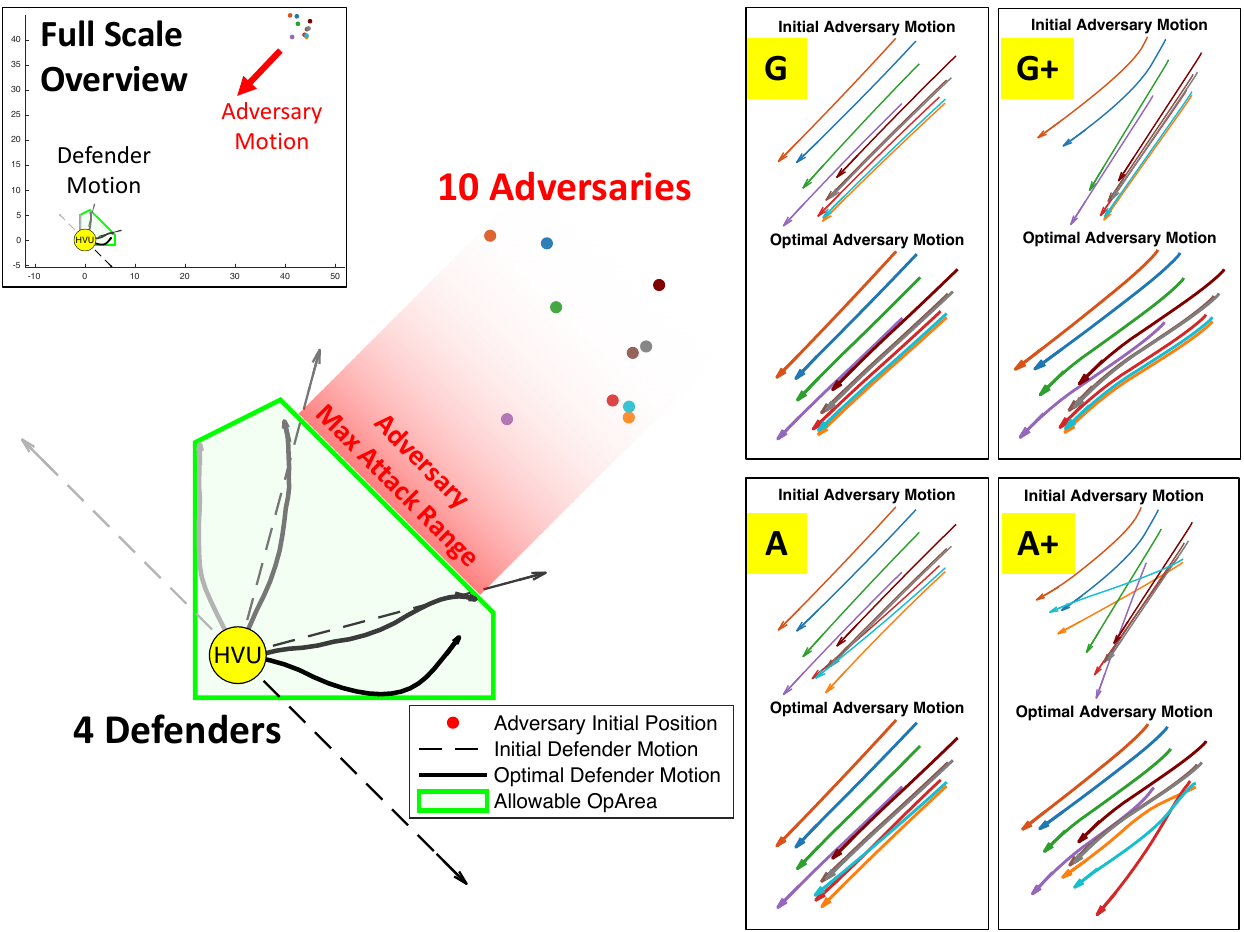}
    \caption{Subplots on right show the optimized adversarial response motions for all possible adversarial tactics (G, G+, A, A+) resulting from a single optimized defender motion.}
    \label{fig:odm_oam}
\end{figure}

As an example, Figure \ref{fig:odm_oam} shows the optimized defender trajectories for the minimum number of optimized defenders needed to meet STP requirements lifted from Figure \ref{fig:odm_dn}, as well as the resulting adversary trajectories for each of the four possible adversarial tactics. Notice that the optimized adversary responses are smoother, and trajectory crossings are minimized. Finally, a critical nuance highlighted in this experiment is that the allowable area constraint was specifically selected to keep defenders clear of the maximum adversarial attack range, taking into account both the maximum adversarial weapon engagement range and their maximum forward travel. The defensive tactical implications of this are significant, and these considerations are equally applicable to civilian applications, ensuring that safety margins are maintained during spatial interactions among numerous non-communicating autonomous agents.

\section{Conclusion}

\subsection{Summary of Main Findings}

This study aimed to develop a robust framework for training a NN classifier for swarm defense systems, focusing on improving NN performance under uncertain operational conditions, optimizing controllable inputs, and minimizing resource expenditure. The proposed methodology has two main components: enriching the swarm dataset to ensure robustness and developing a NN performance optimization framework.

The results demonstrated that using combined datasets encompassing a range of operational conditions—such as varying defender numbers, motions, and noise levels—significantly improved NN flexibility and robustness. Specifically, the combined datasets enabled the NN to perform better across a wide range of operational scenarios, which is crucial in uncertain conditions involving variable defender numbers or environmental factors. The defender number and motion experiments clearly showed the benefits of training the NN across diverse conditions, resulting in effective classification with fewer defenders and adaptability under varying tactical constraints.

The optimal control framework for generating optimal defender motion demonstrated its flexibility, operational effectiveness, and potential to produce optimal defender trajectories that are operationally constrained. Additionally, the framework provided insights into the minimum number of defenders required to achieve specific classification confidence levels, showing a clear potential for resource reduction and improved strategic planning.

In conclusion, this study provides a systematic approach for building robust NN classifiers for swarm defense, optimized through enriched datasets and DOF-focused tuning. These contributions lay the groundwork for adaptable, efficient, and responsive swarm defense strategies in both military and broader autonomous agent contexts.

\subsection{Potential Applications}

The framework presented in this study can be applied across both military and civilian domains wherever typical or significant agent behaviors exist and can be effectively simulated at a macro Newtonian level to generate datasets for robust NN training. Additionally, the optimization aspect of the framework can be applied in any system that has controllable links to NN inputs, allowing for improved performance through systematic tuning of those inputs.

In military swarm defense, the approach can be used for developing defense strategies against autonomous swarms, enhancing the ability to classify and counteract adversary tactics efficiently. For autonomous driving, predicting the behavior of surrounding vehicles at intersections, during lane merging, or in collision avoidance scenarios can improve safety and efficiency, especially in urban environments. Similarly, in traffic management for air and sea, the framework can be applied to regulated airspace and congested waterways to improve routing efficiency and collision prevention, which is crucial for maintaining safe and orderly traffic in these regions. In automated warehouses, where both robots and humans co-exist, predicting worker trajectories can reduce risks, optimize routing efficiency, and increase throughput by anticipating potential conflicts and re-routing as needed.

\subsection{Future Work}
\label{sec:future_work}

Future work can extend this study in several significant directions. An interesting direction would be evaluating the trade-offs between generating more engagement instances for NN training versus using the optimization framework for improved classifier performance. These trade-offs could be expressed in terms of training time or computational resources, and then transformed into a performance metric, providing a means of comparison. Additionally, understanding when to shift computational requirements offline, such as from optimization to dataset enrichment, versus keeping them online for increased flexibility, could prove valuable, especially when classifiers might be deployed on edge computing devices with size, weight, and power restrictions.

Another important direction could compare the performance of other NN models using the optimal defender motion derived from our CNN model. This comparison could determine whether the optimized adversary response yields similar improvements with other NN models, produces worse results, or is entirely incompatible. Such insights would help evaluate the generalizability of the optimization framework. For instance, if it is easier or more cost-effective to develop an optimization using a smaller model (e.g CNN), how will the results scale to a larger, potentially more powerful model (e.g. Transformer)?

These future directions will help in developing more efficient, flexible, and effective swarm defense systems, thereby broadening the range of applications and making the approach even more versatile for both military and civilian contexts.

\section*{Acknowledgment}
This work was supported by the Office of Naval Research Science of Autonomy Program under grant N0001424WX01651.

\bibliographystyle{ieeetr}
\bibliography{0swarm2}

\begin{thebibliography}{10}

\bibitem{balmforth_ukrainian_2023}
T.~Balmforth, ``Ukrainian drone disables {Russian} warship near {Russia}'s {Novorossiysk} port,'' {\em Reuters}, Aug. 2023.

\bibitem{robins-early_ais_2024}
N.~Robins-Early, ``{AI}’s ‘{Oppenheimer} moment’: autonomous weapons enter the battlefield,'' {\em The Guardian}, July 2024.

\bibitem{kallenborn_are_2020}
Z.~Kallenborn, {\em Are drone swarms weapons of mass destruction?}
\newblock The {Counterproliferation} {Papers} {Future} {Warfare} {Series}, no. 60, Maxwell Air Force Base, Alabama, USA: U.S. Air Force Center for Strategic Deterrence Studies, Air University, 2020.

\bibitem{pettyjohn_swarms_2024}
S.~Pettyjohn, M.~Campbell, and H.~Dennis, ``Swarms over the {Strait},'' tech. rep., Center for a New American Security, June 2024.

\bibitem{cummings_artificial_2017}
M.~L. Cummings, ``Artificial {Intelligence} and the {Future} of {Warfare},'' tech. rep., International Security Department and US and the Americas Programme, July 2017.

\bibitem{clark_hicks_2023}
J.~Clark, ``Hicks {Underscores} {U}.{S}. {Innovation} in {Unveiling} {Strategy} to {Counter} {China}'s {Military} {Buildup},'' {\em U.S. Department of Defense}, Aug. 2023.

\bibitem{hepworth_report_2022}
A.~Hepworth, ``Report on {Applied} {Research} {Directions} and {Future} {Opportunities} for {Swarm} {Systems} in {Defence},'' Tech. Rep. No. 11, Australian Army Research Center, 2022.

\bibitem{lin_ethics_2013}
P.~Lin, ``The {Ethics} of {Autonomous} {Cars},'' {\em The Atlantic}, Oct. 2013.

\bibitem{jafary_survey_2018}
B.~Jafary, E.~Rabiei, M.~Diaconeasa, H.~Masoomi, L.~Fiondella, and A.~Mosleh, ``A {Survey} on {Autonomous} {Vehicles} {Interactions} with {Human} and other {Vehicles},'' (Los Angeles, CA), Sept. 2018.

\bibitem{emha_abdillah_implementation_2024}
R.~Emha~Abdillah, H.~Moenaf, L.~Fadullah~Rasyid, S.~Achmad, and R.~Sutoyo, ``Implementation of {Artificial} {Intelligence} on {Air} {Traffic} {Control} - {A} {Systematic} {Literature} {Review},'' in {\em 2024 18th {International} {Conference} on {Ubiquitous} {Information} {Management} and {Communication} ({IMCOM})}, pp.~1--7, Jan. 2024.

\bibitem{soori_artificial_2023}
M.~Soori, B.~Arezoo, and R.~Dastres, ``Artificial intelligence, machine learning and deep learning in advanced robotics, a review,'' {\em Cognitive Robotics}, vol.~3, pp.~54--70, Jan. 2023.

\bibitem{vagale_path_2021}
A.~Vagale, R.~Bye, R.~Oucheikh, O.~Osen, and T.~Fossen, ``Path planning and collision avoidance for autonomous surface vehicles {II}: a comparative study of algorithms,'' {\em Journal of Marine Science and Technology}, Feb. 2021.

\bibitem{wurman_coordinating_2008}
P.~R. Wurman, R.~D'Andrea, and M.~Mountz, ``Coordinating {Hundreds} of {Cooperative}, {Autonomous} {Vehicles} in {Warehouses},'' {\em AI Magazine}, vol.~29, pp.~9--9, Mar. 2008.
\newblock Number: 1.

\bibitem{sanchez_ibanez_path_2021}
J.~R. Sánchez~Ibáñez, C.~Perez-del Pulgar, and A.~Garcia, ``Path {Planning} for {Autonomous} {Mobile} {Robots}: {A} {Review},'' {\em Sensors}, vol.~21, p.~7898, Nov. 2021.

\bibitem{amoo_ai-driven_2024}
O.~Amoo, E.~Sodiya, U.~Umoga, and A.~Atadoga, ``{AI}-driven warehouse automation: {A} comprehensive review of systems,'' {\em GSC Advanced Research and Reviews}, vol.~18, pp.~272--282, Feb. 2024.

\bibitem{walton_optimal_2018}
C.~Walton, P.~Lambrianides, I.~Kaminer, J.~Royset, and Q.~Gong, ``Optimal motion planning in rapid‐fire combat situations with attacker uncertainty,'' {\em Naval Research Logistics}, vol.~65, pp.~101--119, Mar. 2018.

\bibitem{walton_defense_2022}
C.~Walton, I.~Kaminer, Q.~Gong, A.~H. Clark, and T.~Tsatsanifos, ``Defense against {Adversarial} {Swarms} with {Parameter} {Uncertainty},'' {\em Sensors}, vol.~22, p.~4773, Jan. 2022.

\bibitem{tsatsanifos_modeling_2021}
T.~Tsatsanifos, A.~H. Clark, C.~Walton, I.~Kaminer, and Q.~Gong, ``Modeling {Large}-{Scale} {Adversarial} {Swarm} {Engagements} using {Optimal} {Control},'' in {\em 2021 60th {IEEE} {Conference} on {Decision} and {Control}}, pp.~1244--1249, Dec. 2021.
\newblock ISSN: 2576-2370.

\bibitem{gong_partial_2020}
Q.~Gong, W.~Kang, C.~Walton, I.~Kaminer, and H.~Park, ``Partial {Observability} {Analysis} of an {Adversarial} {Swarm} {Model},'' {\em AIAA Journal of Guidance, Control, and Dynamics}, vol.~43, pp.~250--261, Feb. 2020.

\bibitem{peltier_swarm_2024}
D.~W. Peltier, I.~Kaminer, A.~Clark, and M.~Orescanin, ``Swarm {Characteristics} {Classification} {Using} {Neural} {Networks},'' {\em IEEE Transactions on Aerospace and Electronic Systems}, pp.~1--12, 2024.

\bibitem{gao_data_2024}
Z.~Gao, H.~Liu, and L.~Li, ``Data {Augmentation} for {Time}-{Series} {Classification}: {An} {Extensive} {Empirical} {Study} and {Comprehensive} {Survey},'' Aug. 2024.
\newblock arXiv:2310.10060 [cs].

\bibitem{iglesias_data_2023}
G.~Iglesias, E.~Talavera, A.~González-Prieto, A.~Mozo, and S.~Gómez-Canaval, ``Data {Augmentation} techniques in time series domain: a survey and taxonomy,'' {\em Neural Computing and Applications}, vol.~35, pp.~10123--10145, May 2023.

\bibitem{nocedal_numerical_2006}
J.~Nocedal and S.~J. Wright, {\em Numerical optimization}.
\newblock Springer series in operations research and financial engineering, New York, NY: Springer, second edition~ed., 2006.

\bibitem{mathworks_inc_fmincon_2024}
{MathWorks, Inc.}, {\em fmincon: {Constrained} nonlinear minimization}.
\newblock MathWorks, Inc., 2024.

\bibitem{mathworks_inc_importnetworkfromtensorflow_2024}
{MathWorks, Inc.}, {\em {importNetworkFromTensorFlow}: {Import} {TensorFlow} network into {MATLAB}}.
\newblock MathWorks, Inc., 2024.

\end{thebibliography}

\begin{IEEEbiography}
[{\includegraphics[width=1in,height=1.25in,clip,keepaspectratio]{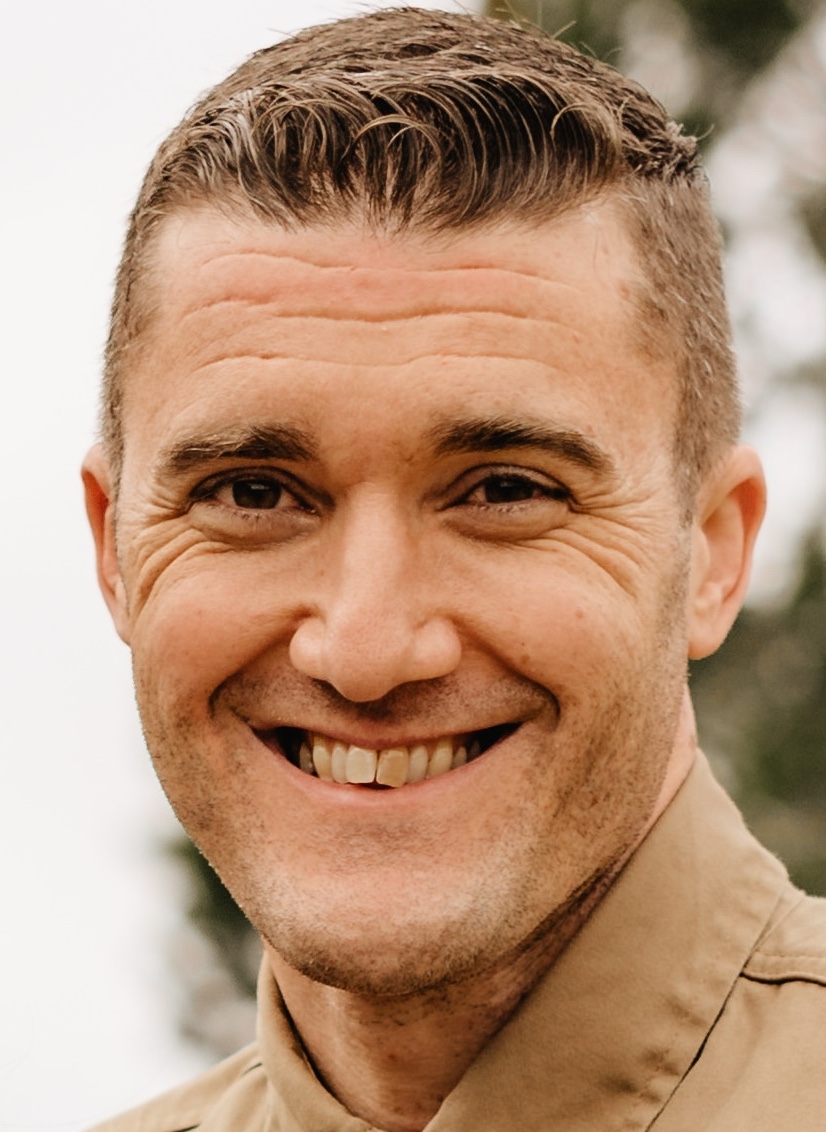}}]
{Donald W. Peltier III}received the B.S. degree in aerospace engineering from The University of Texas at Austin, Austin, TX, USA, in 2006, and the M.S. degree in aeronautical engineering from the Air Force Institute of Technology, Wright-Patterson AFC, OH, USA, in 2007. He is currently working toward the Ph.D. degree in mechanical engineering with a focus in control systems with the Naval Postgraduate School.

His research interests include tasks involving multiple autonomous agents, artificial intelligence assisted control systems, and applications to improve safety and quality of life.
\end{IEEEbiography}

\begin{IEEEbiography}
[{\includegraphics[width=1in,height=1.25in,clip,keepaspectratio]{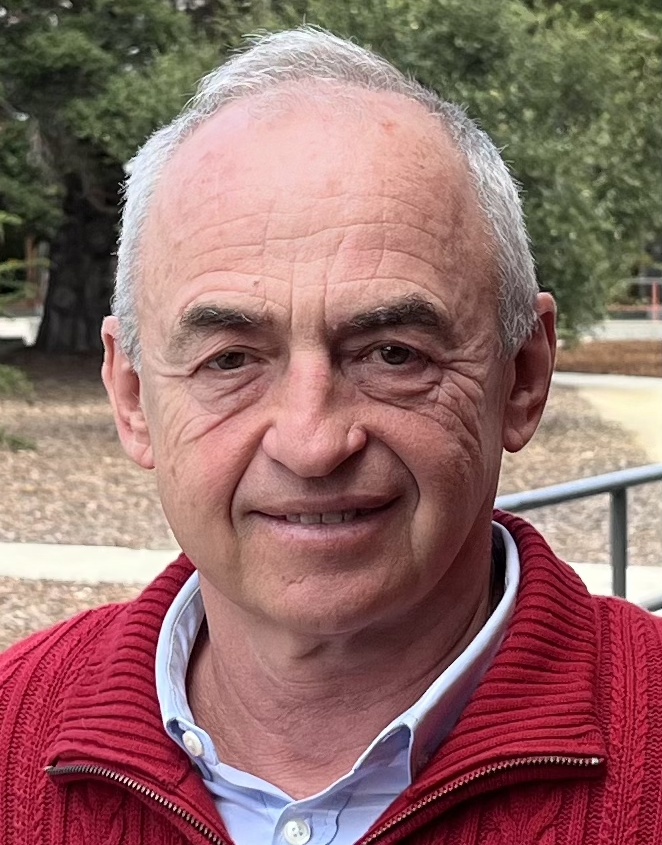}}]
{Isaac Kaminer} received Ph.D. in Electrical Engineering from University of Michigan in 1992. Before that he spent four years working at Boeing Commercial first as a control engineer in 757/767/747-400 Flight Management Computer Group and then as an engineer in Flight Control Research Group. Since 1992 he has been with the Naval Postgraduate School first at the Aeronautics and Astronautics Department and currently at the Department of Mechanical and Aerospace Engineering where he is a Professor. He has a total of over 20 years of experience in development and flight testing of guidance, navigation and control algorithms for both manned and unmanned aircraft. His more recent efforts were focused on development of coordinated control strategies for multiple UAVs and vision based guidance laws for multiple UAVs. Professor Kaminer has co-authored more than two hundred refereed journal and conference publications.
 
He is member of the Institute of Electrical and Electronic Engineers, American Institute of Aeronautics and Astronautics, and Panel Member of the NATO Research Technology Organization, SCI-023 on Unmanned Combat Air Vehicles. He was awarded the NASA Certificate of Recognition for the Creative Development of a Technical Innovation, October 1991; the 1994 NATO Fellowship for Scientific and Technological Exchange, the 1994 Excellence in Research Award, Naval Postgraduate School, the 1995 ASEE/NASA Summer Faculty Fellowship, the 1995 NATO Fellowship for Scientific and Technological Exchange, the 1999 AIAA Outstanding Service Award, the 1999 NPS Menneken Annual Faculty Award for Excellence in Scientific Research, and 2022 IEEE CSS Award for Technical Excellence in Aerospace Control 
\end{IEEEbiography}

\begin{IEEEbiography}
[{\includegraphics[width=1in,height=1.25in,clip,keepaspectratio]{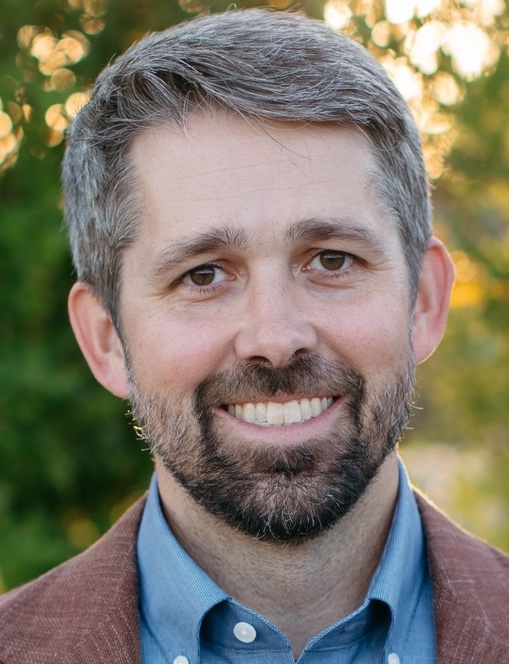}}]
{Abram H. Clark IV} received the B.S. degree in 2006 and the M.S. degree in 2008 in electrical engineering from Texas Tech University, Lubbock, TX, USA, as well as the Ph.D. degree in physics in 2014 from Duke University, Durham, NC, USA.

He was a Postdoctoral Associate from 2014 to 2017 with the Department of Mechanical Engineering and Materials Science, Yale University, New Haven, CT, USA. He is currently an Associate Professor with the Department of Physics, Naval Postgraduate School, where he has been since 2017. His research interests include emergent behavior in soft matter systems (e.g., granular flows) as well as other large, many-body systems (e.g., robot swarms).
\end{IEEEbiography}

\begin{IEEEbiography}
[{\includegraphics[width=1in,height=1.25in,clip,keepaspectratio]{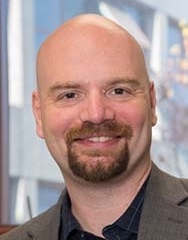}}]
{Marko Orescanin}(Member, IEEE) received a Ph.D. degree in Electrical and Computer Engineering from the University of Illinois Urbana-Champaign, Champaign, IL, USA, in 2010.
Since 2019, he has been an Assistant Professor with the Computer Science Department at the Naval Postgraduate School, Monterey, CA, USA. From 2011 to 2019, he was with Bose Corporation, MA, USA, where he primarily worked on research and advanced development of signal processing and machine learning algorithms for audio and speech enhancement in consumer electronics. He left Bose as a Senior Manager of the AI and Data group with a focus on the consumer electronics business unit. His research interests include signal processing, machine learning, artificial intelligence, Bayesian deep learning, acoustics, passive and active sonar, unmanned vehicles, and remote sensing.
\end{IEEEbiography}

\end{document}